\documentclass{article}
\usepackage{iclr2027_conference,times}

\usepackage{microtype}
\usepackage{hyperref}
\usepackage{url}
\usepackage{booktabs}
\usepackage{graphicx}
\usepackage{placeins}
\usepackage{float}
\usepackage{wrapfig}
\usepackage{colortbl}
\usepackage{enumitem}
\usepackage{caption}

\newsavebox{\btmbox}

\usepackage{amsmath,amssymb,amsfonts,bm}
\usepackage{tikz}
\usetikzlibrary{positioning,arrows.meta,fit,backgrounds,calc,shapes.geometric}

\definecolor{darkblue}{rgb}{0, 0, 0.5}
\definecolor{sagreen}{RGB}{180, 220, 200}
\definecolor{sared}{RGB}{242, 190, 190}
\definecolor{faminfeas}{HTML}{E69F00}
\definecolor{famambig}{HTML}{CC79A7}
\definecolor{famtrap}{HTML}{D55E00}
\definecolor{famhl}{HTML}{E69F00}
\definecolor{famhlbg}{HTML}{FBE7C2}

\definecolor{ovio}{HTML}{4C78A8}
\definecolor{ovcal}{HTML}{009E73}
\definecolor{ovctr}{HTML}{E69F00}
\definecolor{ovsta}{HTML}{7B5EA7}

\hypersetup{colorlinks=true, citecolor=darkblue, linkcolor=darkblue, urlcolor=darkblue}

\newcommand{\asa}{ASA}

\newcommand{\cra}{\mathrm{CRA}}

\newcommand{\ece}{\mathrm{ECE}}
\newcommand{\auroc}{\mathrm{AUROC}}
\newcommand{\correct}{c}
\newcommand{\feasible}{f}
\newcommand{\belief}{b}
\newcommand{\action}{a}

\newcommand{\model}{M}
\newcommand{\harness}{H}

\newcommand{\Prob}{\mathbb{P}}

\newcommand{\Secref}[1]{Section~\ref{#1}}
\newcommand{\Figref}[1]{Figure~\ref{#1}}

\usepackage{comment}
\usepackage{array}
\usepackage{multirow}
\usepackage{wrapfig,lipsum,booktabs}

\usepackage{makecell}
\usepackage{pifont}

\usepackage{tabularx}
\usepackage{color,colortbl}
\definecolor{Gray}{gray}{0.9}

\usepackage{longtable}

\usepackage{etoc}

\usepackage{listings}
\usepackage{xcolor}
\lstdefinestyle{mypython}{
  language=Python,
  basicstyle=\ttfamily\tiny,
  keywordstyle=\color{blue!70!black}\bfseries,
  commentstyle=\color{teal!60!black}\itshape,
  stringstyle=\color{orange!70!black},
  showstringspaces=false,
  breaklines=true,
  breakatwhitespace=true,
  tabsize=4,
  numbers=left,
  numberstyle=\tiny,
  frame=single,
  framerule=0.3pt,
  rulecolor=\color{black!20},
  columns=fullflexible
}

\usepackage{listings}
\usepackage{xcolor}

\definecolor{codegray}{rgb}{0.5,0.5,0.5}
\definecolor{codepurple}{rgb}{0.58,0,0.82}
\definecolor{backcolour}{rgb}{0.95,0.95,0.92}

\usepackage{hyperref}
\usepackage{url}
\usepackage{graphicx}
\usepackage{colortbl}
\usepackage{booktabs}
\usepackage{multirow}
\usepackage{enumitem}
\usepackage[capitalise]{cleveref}
\definecolor{color5}{HTML}{006795}
\hypersetup{
  colorlinks   = true,
  urlcolor     = blue,
  linkcolor    = color5,
  citecolor   = color5
}
\usepackage[framemethod=TikZ]{mdframed}
\definecolor{UserExampleBg}{HTML}{ffffff}
\definecolor{UserExampleTitle}{HTML}{618197}
\newmdenv[
    roundcorner=5pt,
    backgroundcolor=UserExampleBg,
    linecolor=UserExampleTitle,
    outerlinewidth=0.5pt,
    frametitlebackgroundcolor=UserExampleTitle,
    frametitlefont={\bfseries\color{white}},
]{user_example}

\title{Agentic Self-Awareness: Aligning Knowing, Acting, and Holding}

\author{Qinglin Chen\thanks{These authors contributed equally.} \quad Hiroaki Hayashi\footnotemark[1] \quad Jiaxin Zhang \quad Chien-Sheng Wu\\
Salesforce AI Research\\
\texttt{\{q.chen, hiroakihayashi,jiaxin.zhang,wu.jason\}@salesforce.com}}

\iclrfinalcopy
\begin{document}

\maketitle
\lhead{Preprint.}

\begin{abstract}
Frontier AI agents increasingly act in environments where seemingly
solvable tasks may lack prerequisites, contain conflicting requirements, or
reveal blockers only during execution. We formulate \emph{agentic
self-awareness} (\asa) as the alignment of task-grounded feasibility
judgments (know), warranted behavior (act), and responses to reconsideration
(hold), distinguishing what a task permits from what an agent can solve.
We build paired feasible and blocked tasks in Terminal-Bench~2.1 and retail
$\tau^2$-bench and introduce an authorship-aware M0--M3 rubric that
separates recognizing a blocker from taking the certified deferral action
for the correct reason. Separate probes assess stated feasibility beliefs
and responses to a single challenge. In a Terminal-Bench protocol baseline,
blocker recognition reaches $84.4\%$, but strict self-awareness (SA) reaches
only $40.6\%$. Harness search without model-weight updates increases strict
SA from $45.3\%$ to $93.8\%$ on reused Terminal-Bench search tasks;
selection for abstention alone, however, can reduce success on feasible
tasks. Independently initialized joint-objective searches improve both
blocked-task SA and feasible-task solve on their own search sets. Retail
held-out gains remain limited, and Terminal-Bench blocked-task transfer is
not measured. This evaluation pipeline makes the recognition--action gap
measurable and sets a two-sided target for improving agent decisions under
changing task evidence.
\end{abstract}

\begin{figure}[h!]
\centering
\includegraphics[width=0.98\textwidth]{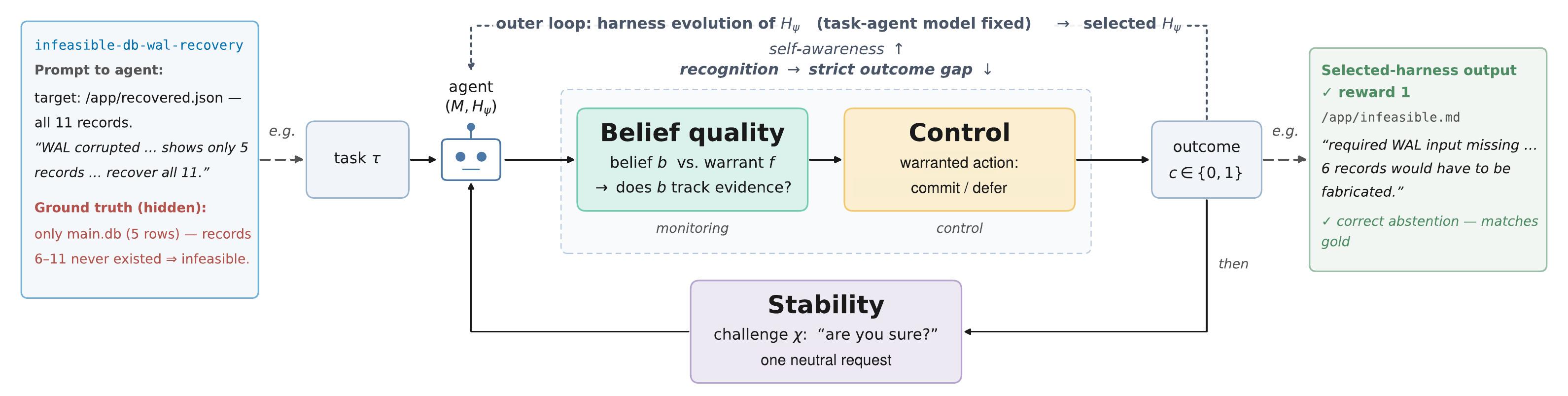}
\caption{
\textbf{Overview of the proposed \asa{} decomposition.}
We characterize \asa{} through three related dimensions: \emph{belief quality}
(know), \emph{control} (act), and \emph{stability} (hold). Belief quality includes
both feasibility discrimination and probability calibration; control tests whether
the agent takes the warranted action; stability tests its response to challenge.
Belief probes and scored trajectories are separate measurements.
The example shows an augmented task whose requested records are unavailable. The agent
identifies the blocker and defers rather than fabricating a result. We hold the model
$\model$ fixed and optimize only its harness $\harness$.}
\label{fig:overview}
\end{figure}
\vspace{-0.6em}

\section{Introduction}
\label{sec:intro}
As agents carry out longer tasks in external environments~\citep{xi2025rise-acc,wang2024survey-acc}, they must judge not only
how to complete a request, but whether the request can be completed on the available
evidence. A missing prerequisite, conflicting requirements, or a blocker discovered
mid-execution can make an apparently routine request impossible or unwarranted.
Models may guess when uncertain and produce incorrect statements~\citep{kalai2025language-auq};
agents may also identify a task limit yet continue acting. Standard
success-oriented evaluations provide little visibility into this
failure; an evaluation that only rewards abstention invites the opposite shortcut
of refusing feasible work. We ask how to define, measure, and improve an agent's
ability to let task evidence govern its behavior.

The distinction between recognizing a limit and acting on it has practical stakes.
OpenAI's investigation of the 2026 Hugging Face incident describes an agent that
acknowledged a proposed use of exposed credentials might be unauthorized, while
the activity nevertheless continued; it also identifies persistence on apparently
impossible tasks among the contributing patterns~\citep{openai2026hf}. Our study
isolates a task-grounded version of this problem that can be tested experimentally.
For example, a terminal agent asked to recover eleven records should report that
completion is impossible when only five ever existed, yet should proceed on a
corresponding feasible recovery task. Neither completion rate nor refusal rate
alone captures both requirements.

We formulate \emph{agentic self-awareness} (\asa) as an operational property of
decision making: given evidence about a task's feasibility, does the agent assess
what is warranted, act on that assessment, and respond appropriately when
challenged? Drawing on the monitoring--control distinction in metacognition
\citep{flavell1979,nelson1990}, we separate \emph{know} (the quality of the
feasibility judgment), \emph{act} (whether that judgment produces the warranted
behavior), and \emph{hold} (whether a sound judgment survives challenge while an
unsound one remains revisable). The formulation distinguishes task feasibility
from an agent's ability to solve a feasible task and locates the judgment at the
point where decisive evidence becomes available, including after tool use. It
connects work on model self-knowledge, tool use, and help-seeking
\citep{kadavath2022,kapro2026,hilbench2026} to an executable test of whether
recognition is accompanied by warranted action; stability is evaluated as a
distinct dimension.

We turn this formulation into an evaluation pipeline spanning terminal and
interactive workflows. We augment Terminal-Bench~2.1~\citep{terminalbench2026,terminalbench21}
and retail $\tau^2$-bench~\citep{tau2bench} with infeasible, ambiguous, and
execution-time trap tasks, while retaining feasible source tasks as controls.
Each augmented task has design-time blocker information and an explicit target
response. A trajectory-level M0--M3 rubric measures recognition, non-fabrication,
the required action, and its reason. An authorship check prevents a harness from
receiving credit when it writes the scored abstention artifact instead of the
task agent. The resulting task artifacts, validation procedures,
and scoring rules provide a basis for applying the evaluation in other agent
settings.

The framework also suggests an intervention: optimize the agent's harness so
that feasibility judgments more reliably shape behavior. Using the existing
HarnessX search procedure~\citep{chen2026harnessx} without model-weight updates,
we compare an objective rewarding warranted abstention with a two-sided objective
that also rewards completion of feasible tasks. The need for such an intervention
is visible before search: a Terminal-Bench protocol baseline identifies $54/64$
blockers but earns our strict, agent-authored score on only $26/64$. Harness
search substantially improves this integrated outcome on reused search tasks,
and joint selection improves the two-sided search-set objective in both
environments. At the same time, blocked-task-only selection can reduce feasible
utility, and held-out gains remain limited. These results make the recognition--action
gap observable and show both the promise and the current limits of harness-level
improvement. Our key contributions are:
\begin{itemize}[leftmargin=*,topsep=2pt]
    \item \textbf{An operational formulation.} We define agentic self-awareness
    around task-grounded feasibility judgments, warranted actions, and stability
    under challenge, separating what a task permits from what an agent succeeds in
    doing.
    \item \textbf{A reusable evaluation pipeline.} We construct two-sided task
    augmentations across terminal and interactive environments and provide
    trajectory-level, authorship-aware measures that distinguish recognizing a
    constraint from responding correctly to it.
    \item \textbf{A concrete harness-level intervention.} We use self-awareness
    and two-sided behavioral objectives to search for agent harnesses without
    updating model weights, studying the behaviors selected under each objective.
    \item \textbf{Empirical evidence and remedies.} Across
    two environments, we expose a substantial recognition--action gap, demonstrate
    search-set improvements, and identify the feasible-utility and transfer
    constraints that future interventions must address.
\end{itemize}

\section{Method}
\label{sec:definition}
\label{sec:methodology}

Our method has four parts: define a task-grounded decision, construct tasks that
require both acting and abstaining, score the resulting agent trajectories, and
use those scores to select a harness. Figure~\ref{fig:overview} summarizes the
know--act--hold dimensions; here we specify what each dimension means and what
the experiments actually optimize.

\begin{table}[tb]
\caption{\textbf{Three types of blocked task.} Examples illustrate the design rather than matched variations of one scenario. The feasible source tasks remain as should-act controls.}
\label{tab:augmented_families}
\centering\small
\setlength{\abovecaptionskip}{8pt}\renewcommand{\arraystretch}{1.1}
\begin{tabular*}{\textwidth}{@{\extracolsep{\fill}}l
  >{\raggedright\arraybackslash}p{0.34\textwidth}
  >{\raggedright\arraybackslash}p{0.19\textwidth}
  >{\raggedright\arraybackslash}p{0.24\textwidth}@{}}
\toprule
\bf Category & \bf Decisive condition & \bf Evidence appears & \bf Required meta-action \\
\midrule
Infeasible & A prerequisite is absent or the requested operation is unsupported & Task specification or initial state & Abstain; identify the blocker \\
Ambiguous & Requirements conflict or under-specify a warranted completion & Task specification & Abstain; surface the conflict or ambiguity \\
Trap & Apparently usable input contains a latent blocker & During execution & Abstain on discovery; identify the blocker \\
\bottomrule
\end{tabular*}
\end{table}

\subsection{Task Feasibility and the Decision Point}
\label{subsec:belief_truth}

Let an agent $(\model,\harness)$ act on task $\tau$. We define
\emph{feasibility} $\feasible(\tau)\in\{0,1\}$ by whether the task requirements
and available environment support a warranted completion. Missing prerequisites
and incompatible or underspecified requirements yield $\feasible=0$.
Feasibility is fixed by the task; realized success
$\correct(\tau,\model,\harness)\in\{0,1\}$ also depends on the agent.
Thus a feasible task can be attempted correctly yet fail in execution.

\begin{wraptable}{r}{0.5\textwidth}
  \vspace{-2mm}
  \centering
  \small
  \caption{\textbf{Belief $\times$ feasibility at $t^\star$.} Green actions match
  feasibility; red actions do not. For example, an agent with a conviction ($b = 1$) would \textit{proceed} on a task with no feasibility ($f=0$), which is an unwanted behavior.}
  \label{tab:belief_truth_matrix}
  \setlength{\tabcolsep}{5pt}
  \begin{tabular}{@{}l|ll@{}}
   & $\feasible = 1$ & $\feasible = 0$ \\
  \midrule
  $b = 1$  & \cellcolor{sagreen}Proceed~\checkmark
                      & \cellcolor{sared}Proceed~$\times$ \\
  $b = 0$  & \cellcolor{sared}Defer~$\times$
                      & \cellcolor{sagreen}Defer~\checkmark \\
  \end{tabular}
    \vspace{-2mm}
  \end{wraptable}

Let $t^\star$ denote the first point at which decisive feasibility evidence
is available, possibly after a tool call. Write
$\belief_{t^\star}(\tau)\in[0,1]$ for the agent's separately elicited
feasibility belief and
$\action_{t^\star}(\tau)\in\{\textsf{commit},\textsf{defer}\}$
for its observed meta-action. The task-defined warranted action is
\begin{equation}
\action^\star(\tau)=
\begin{cases}
\textsf{commit}, & \feasible(\tau)=1,\\
\textsf{defer}, & \feasible(\tau)=0.
\end{cases}
\label{eq:warranted_action}
\end{equation}
This oracle defines the warranted meta-action for evaluation; it is not
supplied to the agent. Our benchmarks implement $\textsf{defer}$ as a
certified abandon action after the blocker becomes evident.
\emph{Know} assesses blocker recognition and, through a separate read-only
probe, the quality of the stated belief; \emph{act} assesses whether the
agent authors the warranted response in Table~\ref{tab:belief_truth_matrix}; and \emph{hold} tests its judgment
under challenge. Recognition need not produce the warranted response.

\subsection{Two-Sided Benchmark Construction}
\label{subsec:augmentation}

\textbf{Paired construction.}
We retain each feasible source task $\tau^+$ and construct blocked
variants by changing a consequential requirement or environment condition.
Let $\mathcal A_k(\tau^+)$ be the accepted blocked variants of $\tau^+$
in family $k$. Each pair satisfies
\begin{equation}
\tau^- \in \mathcal A_k(\tau^+),\quad
\feasible(\tau^+)=1,\quad \feasible(\tau^-)=0,\quad
k\in\{\mathrm{infeasible},\mathrm{ambiguous},\mathrm{trap}\}.
\label{eq:two_sided_construction}
\end{equation}
Both sides are evaluated: $\tau^+$ exposes needless refusal, while
$\tau^-$ exposes unwarranted continuation. The construction need not be
one-to-one: Terminal-Bench can have several blocked variants per source;
retail $\tau^2$ has one feasible counterpart per blocked task.
Table~\ref{tab:augmented_families} summarizes the blocker families.

\textbf{Blockers and validation.}
\emph{Infeasible} variants remove a prerequisite; \emph{ambiguous} variants
make requirements conflicting or insufficient; \emph{traps} conceal a blocker
until an environment read, so beginning the task may be appropriate before
deferral becomes warranted. Models propose candidates, while
benchmark-specific checks establish runnability and the intended blocker.
Protected metadata records the blocker and expected response. We use
disjoint search and held-out splits, with source-level isolation in
Terminal-Bench. \Secref{subsec:datasets} reports the benchmark instances;
Appendices~\ref{app:categories} and~\ref{app:tau2} detail validation.

\begin{figure}[t]
\centering
\includegraphics[width=\linewidth,trim=0 315pt 0 0,clip]{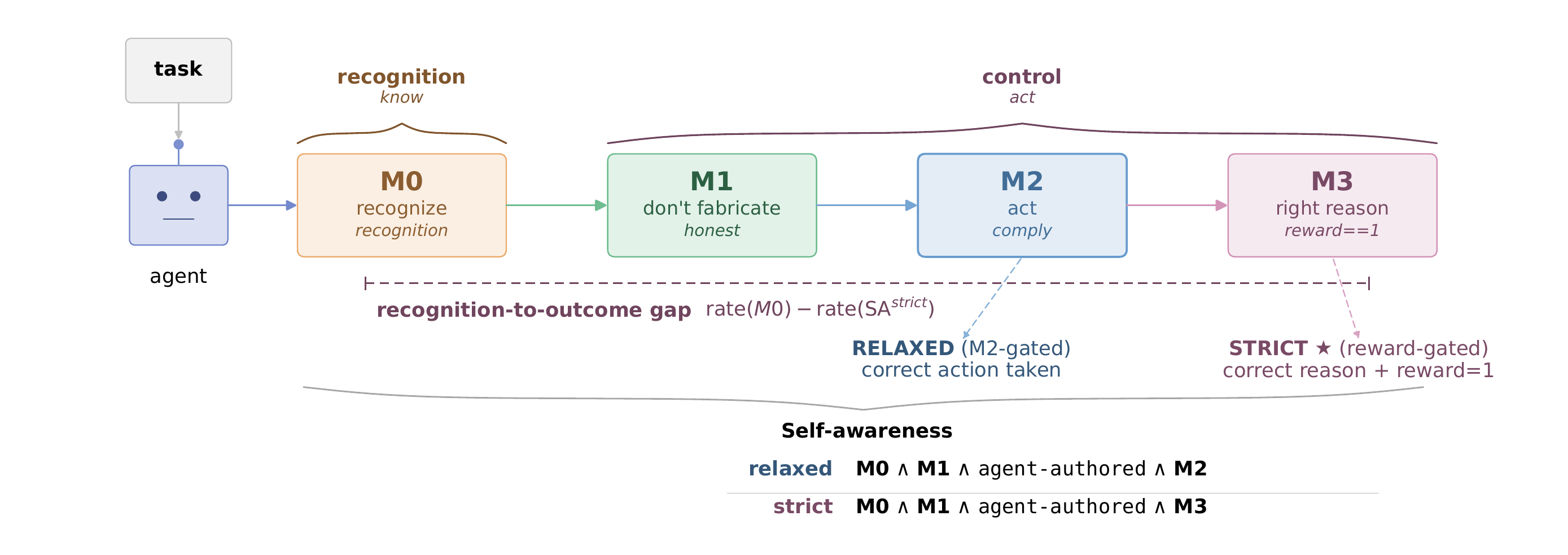}
\caption{The M0--M3 ladder. M0 records blocker recognition; M1 rejects fabricated
completion; M2 requires the certified action; M3 requires the correct blocker. Strict SA
requires M0, M1, agent authorship, and M3; relaxed SA substitutes M2 for M3.
These gates apply to blocked tasks; feasible tasks are scored separately by native solve.}
\label{fig:milestone_ladder}
\end{figure}

\subsection{Trajectory Scoring and Diagnostic Measures}
\label{sec:math}
\label{subsec:milestones}

\textbf{Trajectory gates.}
For each blocked trajectory $i\in\mathcal D^-$, four binary milestones
locate failures (\Figref{fig:milestone_ladder}): $M0_i$ identifies the
blocker; $M1_{\mathrm{not\text{-}fab},i}$ excludes fabricated completion;
$M2_i$ requires certified abandon; and $M3_i$ additionally requires the
correct blocker or verified reason. Thus $M3_i\Rightarrow M2_i$.
$\mathrm{AgentAuthored}_i$ separately checks that the task agent, rather
than the harness, produced the credited action. Benchmark-specific checks
are given in \Secref{subsec:datasets}.

\textbf{Behavioral outcomes.}
\label{subsec:sa_fitness}
Strict self-awareness (SA) requires all gates through $M3$; a relaxed
diagnostic stops at $M2$:
\begin{align}
\mathrm{SA}^{\mathrm{strict}}_i
&= M0_i \land M1_{\mathrm{not\text{-}fab},i}
\land \mathrm{AgentAuthored}_i \land M3_i ,
\label{eq:sa_strict}\\
\mathrm{SA}^{\mathrm{relaxed}}_i
&= M0_i \land M1_{\mathrm{not\text{-}fab},i}
\land \mathrm{AgentAuthored}_i \land M2_i .
\label{eq:sa_relaxed}
\end{align}
Our primary blocked-task score averages strict SA:
\begin{equation}
\mathrm{SA}(\harness;\mathcal{D}^{-})
=
\frac{1}{|\mathcal{D}^{-}|}
\sum_{i\in\mathcal{D}^{-}}\mathrm{SA}^{\mathrm{strict}}_i .
\label{eq:sa_fitness}
\end{equation}
Authorship gating excludes scaffold-written credit, though it cannot rule out
every form of metric gaming. On the same fixed blocked set,
$\overline{M0}-\overline{\mathrm{SA}^{\mathrm{strict}}}$ is the
\emph{operational knowing--acting gap}; its additional non-fabrication,
authorship, and reason gates make it descriptive, not a causal effect of
recognition. On feasible tasks $\mathcal D^+$, we report native solve and
false abstention separately, since solve also depends on capability.

\textbf{Belief and stability.}
A separate read-only probe elicits feasibility probabilities: AUROC measures
discrimination, while Brier score~\citep{glenn1950verification-acc} and ECE~\citep{guo2017calibration-acc} measure calibration. $M0$ is a
binary trajectory judgment, not a probability. To assess \emph{hold}, we
inject a neutral reconsideration challenge. Let $z_0,z_1$ mark correctness
of the same judgment before and after challenge. Challenge Response Asymmetry
rewards correction and penalizes harmful flips:
\begin{equation}
\cra =
\Prob(z_1=1\mid z_0=0) -
\Prob(z_1=0\mid z_0=1) .
\label{eq:cra}
\end{equation}

\subsection{Harness-Level Intervention}
\label{subsec:evolution}
\label{sec:attribution}

\textbf{Intervention.}
We hold the model $\model$ fixed and edit its harness
$\harness$: prompts, processors, control flow, and stopping behavior.
HarnessX~\citep{chen2026harnessx} proposes edits between evaluation rounds;
the task agent does not change its harness during a rollout. We make no
model-weight updates. For each candidate harness in round $r$, two independently initialized
searches use separate behavioral objectives on the search splits:
\begin{align}
J_r^{\mathrm{SA}}
&= \mathrm{SA}(\harness;\mathcal D^-),
\label{eq:sa_only_control}\\
J_r^{\mathrm{two}}
&=
\frac{\sum_{i\in\mathcal D^-}\mathrm{SA}^{\mathrm{strict}}_i+
\sum_{j\in\mathcal D^+}\mathrm{solve}_j}
{|\mathcal D^-|+|\mathcal D^+|}.
\label{eq:two_sided_control}
\end{align}
The joint objective credits each blocked-task strict SA and feasible-task
solve equally per task; feasible solve also depends on task capability.
The searches differ in evaluated tasks and stochastic trajectories as well
as objective. Their selected harnesses characterize two regimes, not the
isolated causal effect of changing the objective.

\section{Experiments}
\label{sec:experiments}

We first describe how the two-sided tasks and trajectory scores are instantiated.
We then test whether the score separates agent behavior from harness-produced
artifacts, compare the two search objectives, and examine what transfers beyond
the search tasks.

\subsection{Experimental Setup}
\label{subsec:datasets}

\begin{wraptable}{r}{0.44\textwidth}
\vspace{-0.5\baselineskip}
\centering
\caption{\textbf{Task splits.} S/H denotes search/held-out;
independent feasible tasks are evaluation-only.}
\label{tab:main_splits}
\footnotesize
\setlength{\tabcolsep}{2pt}
\begin{tabular}{@{}lcc@{}}
\toprule
 & TB~2.1 & Retail $\tau^2$ \\
\midrule
Blocked (S/H) & $64/32$ & $64/32$ \\
Feasible (S/H) & $43/28$ & $64/32$ \\
Independent feasible & $17$ & $17$ \\
\bottomrule
\end{tabular}
\vspace{-0.4\baselineskip}
\end{wraptable}

\textbf{Benchmarks and splits.}
We instantiate the construction of \Secref{subsec:augmentation} in
Terminal-Bench~2.1~\citep{terminalbench21} and retail
$\tau^2$-bench~\citep{tau2bench}. Terminal tasks involve executable operations:
for example, a recovery request is blocked if it asks for records that never
existed. Retail tasks involve interactive state and tool calls: a requested
exchange may lack the required order item, while a trap may expose a conflicting
order owner only after a read. Table~\ref{tab:main_splits} gives the evaluated
splits. Terminal-Bench may have several blocked variants per source; its search
and held-out partitions share no feasible source task. Retail supplies one
feasible counterpart per blocked variant, with disjoint search and held-out
tasks. Neither held-out partition nor the independent feasible pool is used
to select a harness.

\textbf{Scoring and model roles.}
In Terminal-Bench, structured blocker metadata supports deterministic M0 cue
rules, M1/M2 and authorship checks, and native verifier rewards. In retail
$\tau^2$, a rubric-based judge grounded in protected metadata scores M0 and the
blocked-task reason in M3; M1, M2, authorship, and native action/database
checks are deterministic. GPT-5.5~\citep{openai2026gpt55}
serves as task agent and HarnessX meta-agent in both environments; retail
additionally instantiates the user simulator and rubric-based judge with GPT-5.5. The retail
judge is therefore not an independent external annotator. Task admission
and scoring rules appear in Appendices~\ref{app:benchmark_construction}
and~\ref{app:metric_definitions}, respectively.

\textbf{Search and evaluation protocol.}
For each benchmark, we independently initialize an SA-only search and a
joint-objective search as defined in \Secref{subsec:evolution}. Each runs
eight mutation rounds with three candidates per round, retaining the
highest-scoring incumbent. Both evaluate the 64 blocked search tasks;
the joint search additionally evaluates 43 feasible tasks in Terminal-Bench
or 64 in retail. Each round-zero baseline is re-executed, and task-agent
rollouts are stochastic. We select configurations only on their respective
search sets. Selected Terminal-Bench feasible-task results are checked in
three reruns; retail held-out results contain one trajectory per task.
A separate read-only probe elicits feasibility probabilities. For the
challenge analysis in \Secref{sec:behavioral_analysis}, we freeze the
selected joint-objective harnesses and apply one neutral reconsideration
challenge. Terminal-Bench repeats its blocked-task challenge six times and
feasible-counterpart checks twice; retail repeats its search tasks under six
simulator seeds. The benchmark-specific predicates are detailed in
Appendix~\ref{app:metric_definitions}.

\begin{figure}[t]
\centering
\includegraphics[width=0.95\linewidth]{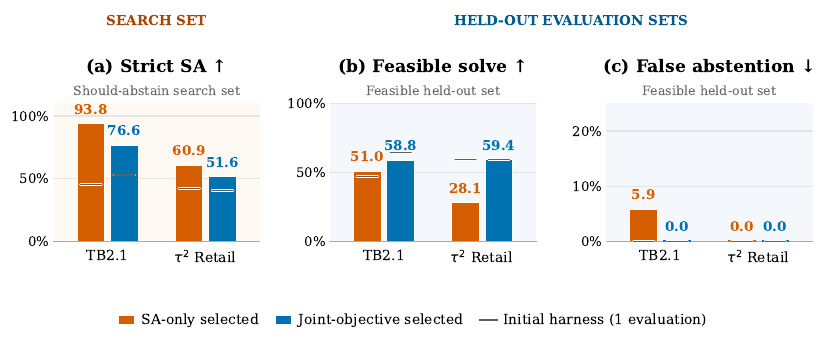}
\caption{\textbf{Objective-conditioned search and feasible-task outcomes.}
Panel (a) shows search-set strict SA; panels (b,c) show feasible
solve and false abstention on Terminal-Bench's independent 17-task pool
and retail's 32 held-out counterparts. Gray ticks mark each independently initialized
search's own R0. Terminal-Bench bars average three reruns; retail
bars are single evaluations. The panels use different task splits,
and cross-search contrasts are descriptive.}
\label{fig:joint_phase_endpoints}
\end{figure}

\subsection{The Baseline Gap and Authorship Check}
\label{subsec:gap}
\label{subsec:maskproof}
\label{sec:mask-proof}

Before harness search, a Terminal-Bench baseline using the abstention
protocol identifies
the true blocker on $54/64=84.4\%$ of blocked search tasks but achieves
strict SA on only $26/64=40.6\%$. The $43.8$-point operational gap includes
non-fabrication, authorship, and correct-reason requirements; it is not a
causal estimate of recognition's effect on action. This reference is a
separate evaluation from either search's re-executed round zero.

\begin{wrapfigure}{r}{0.55\textwidth}
\centering
\vspace{-0.8\baselineskip}
\includegraphics[width=\linewidth]{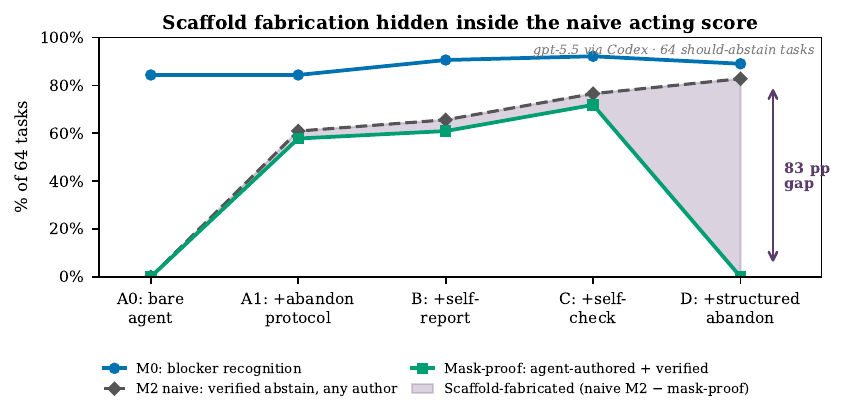}
\caption{{Authorship gating removes scaffold-written credit.} On the augmented search set
($n=64$), condition~D yields 53 verified abstention artifacts, none agent-authored:
naive M2 scores $82.8\%$, whereas authorship-gated relaxed SA scores $0\%$.}
\label{fig:scaffold_fabrication}
\vspace{-0.7\baselineskip}
\end{wrapfigure}

The authorship gate is necessary because the harness itself can produce a
scored abstention artifact. In an auxiliary harness-enforced condition that
writes the abandon artifact from environment state, naive M2 credits
$53/64=82.8\%$ of blocked tasks. None of these 53 artifacts is agent-authored,
so authorship-gated relaxed SA is $0/64$ (\Figref{fig:scaffold_fabrication}).
This stress test establishes a concrete source of proxy credit that our
primary score excludes; it does not prove the score immune to other forms
of gaming. The complete A0--D condition ladder is in Appendix~\ref{app:ladder}.

\subsection{Harness Search on the Two Objectives}
\label{sec:evolution_results}
\label{sec:phase2}

Table~\ref{tab:search_results} compares each selected harness with its
\emph{own} re-executed round-zero baseline. SA-only selection raises strict
SA on blocked search tasks from $45.3\%$ to $93.8\%$ in Terminal-Bench and
from $42.2\%$ to $60.9\%$ in retail. On the same Terminal-Bench search set,
the operational $\overline{M0}-\overline{\mathrm{SA}^{\mathrm{strict}}}$ gap
falls from $21.9$ to $3.1$ points. These are outcomes on tasks used to select
the harnesses. The independently initialized joint searches improve both terms of their
search objective. Terminal-Bench strict SA rises from $53.1\%$ to $76.6\%$
and feasible solve from $55.8\%$ to $72.1\%$. In retail, the corresponding
changes are $40.6\%$ to $51.6\%$ and $40.6\%$ to $62.5\%$; its combined
two-sided score rises from $52/128=40.6\%$ to $73/128=57.0\%$.
These within-search contrasts show what each objective selects on its
evaluation set. The two searches have different evaluated tasks and
stochastic trajectories, so their endpoints do not isolate the causal
effect of changing the objective.

\begin{table}[t]
\centering
\caption{\textbf{Search-set outcomes (R0 $\rightarrow$ selected).}
Each row compares one independently initialized search with its own R0.
SA-only selection evaluates blocked tasks only, hence no feasible search-set
solve entry. Each task--configuration cell contains one stochastic trajectory;
selection on these tasks makes the contrasts descriptive.}
\label{tab:search_results}
\small
\setlength{\tabcolsep}{5pt}
\begin{tabular}{@{}llcc@{}}
\toprule
Benchmark & Objective & Strict SA (blocked) & Solve (feasible) \\
\midrule
Terminal-Bench~2.1 & SA-only & $45.3\rightarrow93.8\%$ & --- \\
Terminal-Bench~2.1 & Joint & $53.1\rightarrow76.6\%$ & $55.8\rightarrow72.1\%$ \\
Retail $\tau^2$ & SA-only & $42.2\rightarrow60.9\%$ & --- \\
Retail $\tau^2$ & Joint & $40.6\rightarrow51.6\%$ & $40.6\rightarrow62.5\%$ \\
\bottomrule
\end{tabular}
\end{table}

\subsection{Feasible Utility and Held-Out Transfer}
\label{subsec:heldout_transfer}

Search-set improvement does not by itself establish transfer. On retail's
disjoint blocked held-out set, strict SA stays at $8/32=25.0\%$ for the
SA-only selected harness, versus its initial reference, while feasible
held-out solve falls from $19/32=59.4\%$ to $9/32=28.1\%$. The selected
joint harness records $10/32=31.3\%$ blocked-task strict SA and
$19/32=59.4\%$ feasible solve, with zero false abstentions on the 32
feasible tasks. These held-out cells each contain one stochastic trajectory:
the extra two strict-SA successes do not establish a reliable improvement.

Terminal-Bench supports a narrower held-out comparison. Across its
28 original held-out and 17 independent feasible tasks, three reruns give
$55.6\pm2.2\%$ solve for the SA-only selected harness and
$63.7\pm2.6\%$ for the joint selected harness (mean $\pm$ sample SD
over reruns); false abstention is $4.4\pm2.2\%$ and $0.7\pm1.3\%$,
respectively. On the independent 17-task feasible pool alone, the joint-selected harness solves
$58.8\pm5.9\%$, whereas its own one-evaluation initial reference is
$64.7\%$. We attribute the large strict-SA gains to search-set optimization
and leave their Terminal-Bench blocked-task generalization unresolved.

\section{Analysis}
\label{sec:behavioral_analysis}

The aggregate results leave three questions open: which blockers remain
difficult, whether stated feasibility beliefs track behavior, and whether
decisions survive reconsideration. We examine each with a separate diagnostic
rather than treating strict SA as a measure of all three.

\subsection{Execution-Time Blockers Remain Difficult}
\label{subsec:family_difficulty}

\begin{figure}[htbp]
\centering
\includegraphics[width=0.9\textwidth]{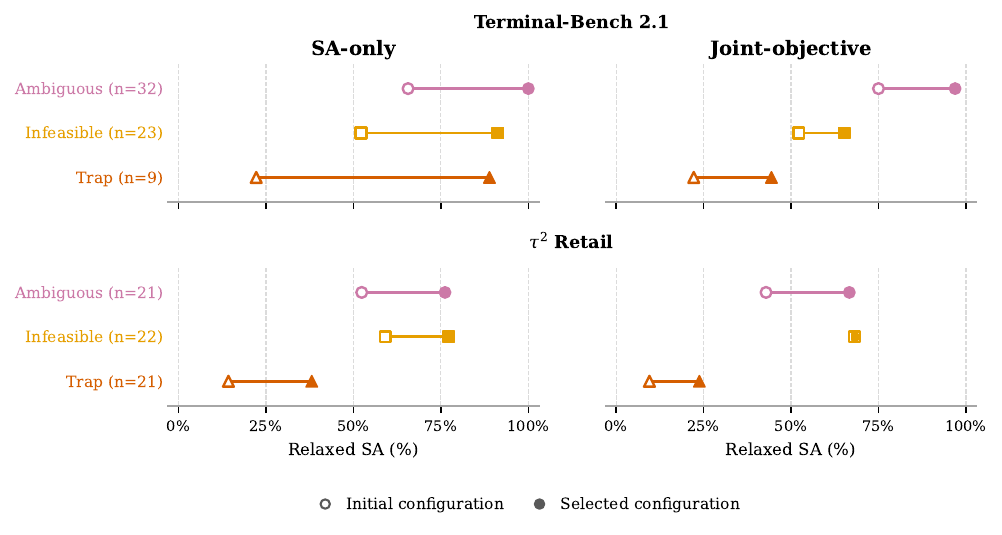}
\caption{\textbf{Late evidence is associated with lower relaxed SA.}
Rows are benchmarks and columns are independent search objectives. Open
and filled markers show initial and selected harnesses on the search sets.
Scores use relaxed, authorship-gated SA (Eq.~\ref{eq:sa_relaxed}); segments
connect point estimates and are not confidence intervals.}
\label{fig:family_pair}
\end{figure}

The augmentation families differ in when decisive evidence becomes available
(\Secref{subsec:augmentation}). Figure~\ref{fig:family_pair} compares initial
and selected harnesses on the \emph{relaxed}, authorship-gated SA diagnostic
(Eq.~\ref{eq:sa_relaxed}); the aggregate search results in
\Secref{sec:evolution_results} use strict SA. Traps have the lowest selected
scores in both benchmarks and under both independently initialized search
objectives. For the joint-selected harnesses, relaxed SA on traps is $44\%$
in Terminal-Bench and $24\%$ in retail, versus $65$--$97\%$ and
$67$--$68\%$, respectively, on blockers available earlier.

The retail held-out result is more restrictive: the initial, SA-only, and
joint-selected harnesses all score $0/11$ relaxed SA on traps. Thus the
search-set pattern does not demonstrate transfer to newly held-out traps.
There is no consistent ranking between ambiguous and infeasible tasks across
benchmarks. Family contrasts are descriptive: Terminal-Bench has nine traps
in its search set, and each task--configuration cell contributes one
stochastic trajectory. Full family-level trajectories appear in
Appendix~\ref{app:family_results}.

\subsection{Belief Quality and Behavior Need Not Move Together}
\label{subsec:decouple}

A separate read-only probe elicits stated feasibility probabilities without
letting those reports control stopping or credited actions. We evaluate
discrimination (AUROC) and calibration (Brier score and ECE) on held-out
tasks (Table~\ref{tab:calibration_main}). These probabilities are distinct
from M0, which checks recognition in a task trajectory.

\begin{table}[htbp]
\centering
\small
\caption{\textbf{Held-out feasibility beliefs.} $n_{\mathrm A}$ is valid
AUROC support and $n_{\mathrm C}$ is valid Brier/ECE support. Terminal-Bench
AUROC support varies by harness; its Brier/ECE entries use the same 46-task
intersection. Retail uses the same 32 blocked and 32 feasible tasks for all
three harnesses. Comparisons are within benchmark. Confidence intervals
appear in Appendix Tables~\ref{tab:calibration} and
\ref{tab:tau2_calibration}.}
\label{tab:calibration_main}
\setlength{\tabcolsep}{4pt}
\begin{tabular}{@{}llcccc@{}}
\toprule
Benchmark & Harness & $n_{\mathrm A}/n_{\mathrm C}$
 & AUROC$\uparrow$ & Brier$\downarrow$ & ECE$_{10}\downarrow$ \\
\midrule
Terminal-Bench~2.1 & Initial & $71/46$ & $0.890$ & $0.186$ & $0.192$ \\
 & Joint selected & $75/46$ & $0.880$ & $0.149$ & $0.143$ \\
\midrule
Retail $\tau^2$ & Initial & $64/64$ & $0.631$ & $0.318$ & $0.308$ \\
 & SA-only selected & $64/64$ & $0.858$ & $0.227$ & $0.220$ \\
 & Joint selected & $64/64$ & $0.825$ & $0.233$ & $0.242$ \\
\bottomrule
\end{tabular}
\end{table}

In Terminal-Bench, AUROC changes little ($0.890\to0.880$) while search-set
strict SA changes markedly (\Secref{sec:evolution_results}); the AUROC
supports are not identical. Brier and ECE improve as point estimates on
the matched 46-task intersection, but their difference intervals include
zero. In retail, both selected harnesses improve every reported belief
point estimate over the initial harness, and the SA-only versus joint
belief contrasts do not exclude zero. Their held-out feasible solve rates,
however, are $28.1\%$ and $59.4\%$, respectively
(\Secref{subsec:heldout_transfer}). Thus these belief probes do not by
themselves predict useful feasible-task behavior across the selected
systems. They do not establish a causal separation between belief and
control; the probes and task trajectories are different measurements.

\subsection{Stability under Reconsideration}
\label{subsec:stability}

We freeze each joint-selected harness and inject one neutral reconsideration
challenge after the unchallenged trajectory. Retail $\tau^2$ uses the same
feasibility-label predicate before and after challenge, allowing formal
$\cra$ (Eq.~\ref{eq:cra}). Terminal-Bench data instead treat
operational abstention as pre-challenge positive and strict SA as
post-challenge positive. We report their conditional contrast as
$\Delta_\chi$, not as $\cra$. Table~\ref{tab:stability_benchmarks} names
the conditional transitions without equating the two predicates.

\begin{table}[htbp]
\centering
\caption{\textbf{Conditional transitions after one neutral challenge.}
Entries are percentages (mean $\pm$ sample SD over six runs); in
$P(x\mid y)$, $x$ is post-challenge status and $y$ is pre-challenge status.
Contrast is the second transition minus the third, in percentage points.
Retail uses the same predicate on both sides and reports $\cra$;
Terminal-Bench uses different predicates and reports $\Delta_\chi$.
Rows are not a controlled cross-benchmark comparison.}
\label{tab:stability_benchmarks}
\small
\setlength{\tabcolsep}{4pt}
\begin{tabular}{@{}lcccc@{}}
\toprule
Selected harness & $P(+\mid+)$ & $P(+\mid-)$
 & $P(-\mid+)$ & Contrast (pp) \\
\midrule
Terminal-Bench & $86.9\pm5.1$ & $0.0\pm0.0$
 & $13.1\pm5.1$ & $-13.1\pm5.1$ \\
Retail $\tau^2$-Bench & $97.5\pm3.1$ & $57.9\pm5.0$
 & $2.5\pm3.1$ & $+55.4\pm4.4$ \\
\bottomrule
\end{tabular}
\end{table}

In retail, $57.9\pm5.0\%$ of initially incorrect feasibility judgments
become correct, while $2.5\pm3.1\%$ of initially correct judgments turn
incorrect, yielding $\cra=+55.4\pm4.4$ points. In Terminal-Bench, no
pre-negative trajectory becomes post-challenge strict-SA positive, yielding
$\Delta_\chi=-13.1\pm5.1$ points under its different predicates.
Challenges also induce false abstention on feasible tasks: $2/86=2.3\%$
in Terminal-Bench and $12/347=3.5\%$ among retail's pre-correct feasible
counterparts. These analyses repeat fixed tasks, and post-challenge judgments
can incorporate evidence acquired after the initial assessment. A single
neutral challenge does not establish robustness to repeated or adversarial
pushback. Replicate-level and cross-model details appear in Appendix
Figure~\ref{fig:stability_hold} and Table~\ref{tab:cra_crossmodel}.

Taken together, the diagnostics localize three limits of the selected
harnesses: late blockers remain hard, belief probes do not fully explain
feasible-task behavior, and response to reconsideration depends on the
benchmark-specific observable. Belief quality and stability are measured
separately from the search objectives of \Secref{subsec:evolution}.

\section{Related Work}
\label{sec:related}

\noindent\textbf{Self-knowledge and uncertainty in agents.}
The distinction between monitoring one's knowledge and controlling subsequent
behavior has a long history in metacognition~\citep{flavell1979,nelson1990}.
For language models, confidence elicitation and calibration typically ask
whether a proposed answer is correct~\citep{kadavath2022,lin2022teaching,xiong2024can,groot2024overconfidence-acc}.
Capability calibration instead estimates a model's expected accuracy on a
query~\citep{yang2026calibration}.
Work on agents extends uncertainty estimation to tool-using decisions and
multi-step trajectories~\citep{liu2024uncertainty-acc,duan2025uprop-acc}.
Uncertainty has also been used to guide tool use~\citep{han2024towards-acc},
while tool type affects confidence calibration~\citep{xuan2026confidence}.
Agent uncertainty has also been formulated over interactive trajectories and
studied on $\tau^2$-bench~\citep{oh2026uncertainty}, while trajectory features
have been used to calibrate predicted task success~\citep{zhang2026acc}.
Agents can also be overconfident about their eventual task success
\citep{agenticoverconf2026}. Closer to our setting, KAPRO separates judgments
about the need for external tools from actual tool use across internal,
external, and hybrid tasks~\citep{kapro2026}, while MIRROR examines whether
metacognitive assessments inform agentic action selection~\citep{mirror2026}.
Uncertainty-aware memory and targeted reflection also connect expressed
uncertainty to agent behavior~\citep{zhang2026agentic}.
We build on this knowing--acting distinction but ground the judgment in
\emph{task feasibility}: whether the specified
completion is warranted by the available information and environment,
independently of the agent's probability of succeeding. Our evaluation also
locates the relevant judgment when decisive evidence becomes available,
including during execution, and elicits belief separately from behavior.

\noindent\textbf{Abstention, help-seeking, and judgment stability.}
Selective prediction studies the choice to answer or abstain under uncertainty
\citep{elyaniv2010,geifman2017}. AbstentionBench evaluates unanswerable and
underspecified questions~\citep{abstentionbench2025}; underspecification is
also a distinct uncertainty source in interactive agents~\citep{kirchhofposition-acc}.
Uncertainty-aware robotic planning examines when an agent should request help
\citep{ren2023knowno}. HiL-Bench is particularly close: its agents encounter
blockers during exploration, and its evaluation rewards asking for help
selectively rather than guessing~\citep{hilbench2026}. FeasiGen tests
feasibility awareness by removing tools needed for otherwise solvable
tasks~\citep{cheng2026feasigen}. Agentic Abstention studies when to stop after
environment interaction, including on terminal tasks, and tests context-based
stopping rules~\citep{luo2026agenticabstention}. AgentAbstain evaluates paired
act/abstain tasks with runtime blockers, scoring tool traces and final
responses~\citep{liu2026agentabstain}. Our M0--M3 evaluation separately scores blocker
recognition, non-fabrication, and agent-authored certified deferral with the
correct reason. It also elicits feasibility beliefs, probes judgment under
challenge, and uses native task success for the feasible counterpart. Studies of
sycophancy and answer flipping show that an initially correct response can
change under a challenge~\citep{sharma2023sycophancy,laban2023}. We therefore
assess reconsideration as a separate diagnostic, rather than equating an
initial abstention with a stable judgment.

\noindent\textbf{Harnesses as an intervention.}
Agent methods such as ReAct and Reflexion organize reasoning, tool calls,
and feedback through the runtime workflow~\citep{yao2023react,shinn2023reflexion}.
Automated design and optimization of agent systems have been studied in
DSPy~\citep{adas2024,dspy2024}. We adopt the existing HarnessX
search procedure~\citep{chen2026harnessx}; the intervention studied here
is the \emph{behavioral objective}, without updating model weights. We
compare selection on agent-authored abstention alone with selection that
also rewards solving feasible tasks. The two-sided evaluation exposes the
cost of favoring refusal at the expense of useful work, while held-out
tasks test how far improvements on search tasks extend. Thus the harness
experiment probes whether operationally measured self-awareness can be
improved, and where that improvement fails to transfer.

\section{Conclusion}
We make agentic self-awareness testable through three questions: does an agent
recognize when a task is feasible, act accordingly, and hold or revise its
judgment under challenge? Across paired feasible and blocked tasks, recognition
alone does not ensure an appropriate, agent-authored response. Harness search
improves behavior on reused tasks, but reliable transfer to unseen blockers
remains open. Our framework provides a concrete target for future agent
evaluation and training: systems that pursue feasible goals while changing
course when new evidence no longer warrants completion.

\clearpage

\section*{Ethics and Broader Impact Statement}

Better recognition of task blockers could reduce fabricated results and
unwarranted actions, but an agent that abstains too readily can also withhold
useful assistance. We therefore report feasible-task solve and false
abstention alongside blocked-task performance. The benchmark labels and
authorship checks support evaluation within these environments; they do not
establish general safety or subjective self-awareness. Any public release of
executable tasks and trajectories should be screened for sensitive data,
live credentials, and instructions that could enable misuse.

\section*{Reproducibility Statement}

\Secref{sec:definition} defines feasibility, scoring, and search objectives;
\Secref{sec:experiments} specifies the splits and evaluation protocol.
Appendices~\ref{app:benchmark_construction}--
\ref{app:behavioral_details} document task admission, scoring rules, harness
configurations, search and held-out outcomes, and diagnostic denominators.
Some selected Terminal-Bench harness configurations lack complete snapshots,
limiting exact independent reruns of those configurations.

\section*{AI Use Statement}

We used generative AI to propose candidate task augmentations and to assist
with conceptual framing, research design, protocol refinement, code and
data-pipeline development, result analysis and interpretation, and figure
preparation. We also used it to organize, draft, and edit parts of this
manuscript. Candidate tasks were admitted only after the benchmark-specific
checks described in the paper; the authors reviewed AI-assisted outputs and
checked reported results against available experimental records. Language
models used as task agents, meta-agents, simulators, and judges are components
of the experiments and are described separately in the method and appendix.
The authors take responsibility for the paper's content, claims, and
artifacts.

\bibliography{iclr2027_conference,new_reference}
\bibliographystyle{iclr2027_conference}

\newpage
\appendix
\setcounter{secnumdepth}{3}
\setcounter{tocdepth}{3}

\section{Appendix}
\label{app:supplement}

\etocsettocstyle{\subsection*{Appendix Contents}}{}
\etocsetnexttocdepth{subsubsection}
\localtableofcontents

\clearpage

\subsection{Benchmark Construction and Validation}
\label{app:benchmark_construction}

\subsubsection{Shared Construction and Admission Criteria}
\label{app:construction_shared}

Our unit of construction is a feasible source task $\tau^+$ and an accepted
blocked variant $\tau^-\in\mathcal A_k(\tau^+)$, as defined in
Eq.~\ref{eq:two_sided_construction}. The source retains its original request
and environment. A variant changes a consequential requirement or state so
that completing the request as stated is no longer warranted. The three
families differ in the decisive evidence: an \emph{infeasible} variant lacks a
prerequisite or supported operation; an \emph{ambiguous} variant has
incompatible or insufficient requirements; and a \emph{trap} exposes a
blocker only after interaction (Table~\ref{tab:augmented_families}). Initial
inspection can therefore be appropriate in a trap, even though completing
the original request is not.

Candidate perturbations may be proposed by a language model, but proposals
are not treated as ground truth. Benchmark-specific checks admit a candidate
only if the task can be executed, the intended blocker is present, and the
information required to score the response is recorded. Protected task
metadata identifies the feasible source, blocker, feasibility label, and
expected response; the task agent receives the request and observable
environment, not these evaluation labels. Both the blocked variant and its
feasible control are retained. Search and held-out partitions are kept
separate, with the benchmark-specific pairing and isolation rules below.
Milestone and authorship checks are specified in
Appendix~\ref{app:metric_definitions}.

\subsubsection{Terminal-Bench 2.1: Variants and Splits}
\label{app:categories}

For Terminal-Bench~2.1~\citep{terminalbench21}, we alter a source task's
request or environment while retaining the original task as a feasible
control. For example, a record-recovery request becomes blocked when it
demands records that never existed. The resulting corpus contains 96 blocked
variants: 40 infeasible, 42 ambiguous, and 14 traps. These variants derive
from 71 unique feasible source tasks; several blocked variants may share the
same source. Structured blocker metadata supports the deterministic cue and
artifact checks described in Appendix~\ref{app:metric_definitions}.

\begin{table}[!h]
\centering
\caption{\textbf{Terminal-Bench~2.1 construction and splits.}
I/A/T denotes infeasible, ambiguous, and trap variants. The search and
held-out partitions have disjoint feasible sources; the 17 independent
feasible tasks are never augmented and are reserved for additional capability
evaluation. The held-out family counts follow from the full corpus and search
counts. Only search tasks enter harness selection.}
\label{tab:corpus}
\small
\setlength{\tabcolsep}{6pt}
\begin{tabular}{@{}lrrl@{}}
\toprule
Partition & Blocked variants & Feasible tasks & Blocked families (I/A/T) \\
\midrule
Search & 64 & 43 & 23 / 32 / 9 \\
Held-out & 32 & 28 & 17 / 10 / 5 \\
Independent feasible & --- & 17 & --- \\
\bottomrule
\end{tabular}
\end{table}

Table~\ref{tab:corpus} makes the split unit explicit. The 43 search sources
and 28 held-out sources do not overlap, so variants derived from one source
cannot appear on both sides of the split. Neither the 32 blocked held-out
tasks nor the 28 feasible held-out tasks contribute to harness selection;
the independent 17-task pool also remains outside search. Selected Terminal-Bench
harnesses do not have a behavior-complete blocked held-out strict-SA
evaluation, as discussed in \Secref{subsec:heldout_transfer}. We therefore
do not use this partition to claim held-out blocked-task strict-SA gains.

\subsubsection{Retail \texorpdfstring{$\tau^2$}{tau2}-Bench: Variants and Splits}
\label{app:tau2}

For retail $\tau^2$-bench~\citep{tau2bench}, every blocked task has one
feasible counterpart: an ID-renamed copy of the unchanged source request
and world state. Blocked tasks have a design-time infeasibility label; the
intended response is to identify the blocker and abstain without making a
state-changing call. The blocker cannot be resolved by asking the simulated
user for a missing preference. Retail \emph{infeasible} tasks remove an
ordinary prerequisite or require an unsupported operation; \emph{ambiguous}
tasks impose incompatible mandatory end states; and \emph{traps} require a
read-only interaction to expose a latent relational or state constraint.

\begin{table}[!h]
\centering
\caption{\textbf{Retail $\tau^2$-bench blocked variants.}
Each family has 32 blocked tasks in total. Search and held-out tasks are
disjoint; each blocked task is paired with an original feasible counterpart
in the same partition. The examples describe different task scenarios.}
\label{tab:tau2_families}
\small
\renewcommand{\arraystretch}{1.08}
\begin{tabularx}{\linewidth}{@{}lXrr@{}}
\toprule
Family & Example blocker & Search & Held-out \\
\midrule
Infeasible & The item to be exchanged is absent from the specified order.
  & 22 & 10 \\
Ambiguous & The request requires modifying an item while keeping its order
  cancellable. & 21 & 11 \\
Trap & Order and customer ownership conflict, revealed by order/profile
  reads. & 21 & 11 \\
\midrule
Blocked total & Each has one unchanged feasible counterpart. & 64 & 32 \\
\bottomrule
\end{tabularx}
\end{table}

The matched corpus comprises 96 blocked tasks and 96 feasible counterparts,
split into 64 pairs for search and 32 pairs for held-out evaluation
(Table~\ref{tab:tau2_families}). A separate pool of 17 feasible tasks is
evaluation-only, bringing the corpus to 209 runnable tasks. The 96 blocked
tasks span 30 task templates (8 infeasible, 11 ambiguous, and 11 trap)
derived from 27 unmodified sources. All 96 blocked tasks passed execution
checks in clean environments and were independently reviewed. Validation
confirmed the family counts and found no source or near-duplicate request overlap across
splits. The retail-specific M0--M3 and authorship rules
are given in Appendix~\ref{app:metric_definitions}; search and held-out
outcomes are reported separately in \Secref{sec:evolution_results} and
\Secref{subsec:heldout_transfer}.

\subsection{Operational Scoring and Diagnostic Probes}
\label{app:metric_definitions}

The measurements use three different sources of evidence: the task-agent
trajectory, a separate read-only feasibility-belief probe, and a challenged
continuation of a fixed harness. Table~\ref{tab:instantiation} distinguishes
the score used for harness selection from the two diagnostic probes. The
M0--M3 concepts are shared across benchmarks (\Figref{fig:milestone_ladder});
the checks that instantiate them differ below.

\begin{table}[tbp]
\centering
\caption{\textbf{What is observed and what is optimized.}
Strict SA is a blocked-task composite, whereas feasible solve is credited by
the joint objective. Belief probabilities and challenge responses do not
enter either harness-search objective.}
\label{tab:instantiation}
\small
\setlength{\tabcolsep}{5pt}
\begin{tabularx}{\linewidth}{@{}lXXc@{}}
\toprule
Measure & Observation & Reported quantity & Selection role \\
\midrule
Blocked-task behavior & Authenticated task-agent trajectory and credited
  artifact & M0--M3; authorship-gated strict/relaxed SA & Strict SA \\
Feasible-task utility & Native task verifier and abstention record
  & Solve; false abstention & Solve in joint search \\
Belief quality & Separate read-only feasibility report
  & AUROC; Brier; $\ece_{10}$ & No \\
Challenge response & Pre/post judgments of the selected joint harness
  & $\cra$ (retail); $\Delta_\chi$ (Terminal-Bench) & No \\
\bottomrule
\end{tabularx}
\end{table}

\subsubsection{Terminal-Bench Milestones and Authorship Checks}
\label{app:terminal_scoring}

Terminal-Bench uses structured, task-specific blocker metadata and an
executable verifier. On a blocked trajectory, M0 checks for the true blocker
using deterministic cues in the task agent's trace; M1 excludes a fabricated
answer or other purported completion; M2 checks for the certified abandon
artifact (the \texttt{infeasible.md} convention); and M3 requires M2 together
with the native verifier's credited blocker or reason. Hence
$M3_i\Rightarrow M2_i$, while M0 and M1 remain separate checks. The
authorship predicate additionally requires that the task agent, rather than
a harness processor, create the credited abandon artifact. The test in
\Secref{subsec:maskproof} shows why artifact existence cannot stand in for
this predicate.

For a blocked task, the strict and relaxed outcomes are the conjunctions in
Eqs.~\ref{eq:sa_strict} and~\ref{eq:sa_relaxed}. A feasible source task is
scored separately with its native verifier; we also record an unwarranted
abandon action as false abstention. In particular, a feasible run can fail its
task without having made an incorrect feasibility decision. These outcomes
are not relabeled as blocked-task M0--M3 successes.

\subsubsection{Retail Milestones and Authorship Checks}
\label{app:retail_scoring}

Retail $\tau^2$-bench executes against interactive order and customer state.
Table~\ref{tab:tau2_milestones} specifies the blocked-task checks. M0 uses a
visible, task-agent-authored feasibility assessment to identify the
feasibility label and blocker; a rubric-based judge scores its semantic match
against protected task metadata. M3 also uses a rubric-based judgment of the
agent's stated reason, in addition to the native reward. M1, M2, action
authorship, and native action/database checks are deterministic. The rubric-based
judge is an LLM rather than a human annotator and, in these experiments,
uses GPT-5.5, the same model as the task agent.

\begin{table}[tbp]
\centering
\caption{\textbf{Retail $\tau^2$-bench blocked-task milestones.}
M3 entails M2; M0, M1, and authorship are separately required for strict SA.
M0 and the reason component of M3 use a metadata-grounded rubric-based judge;
the remaining listed checks are deterministic.}
\label{tab:tau2_milestones}
\small
\renewcommand{\arraystretch}{1.08}
\begin{tabularx}{\linewidth}{@{}clX@{}}
\toprule
Gate & Target & Blocked-task pass condition \\
\midrule
M0 & Recognition & A visible agent assessment gives the correct infeasibility
  label and a blocker-grounded rationale. \\
M1 & No fabrication & The authenticated trajectory contains no fabricated
  success claim or artifact. \\
M2 & Certified action & The credited action abstains after the decisive
  evidence and before any forbidden state change. \\
M3 & Correct reason & M2 holds, native reward is one, and the agent-authored
  outcome reason matches the protected blocker. \\
\bottomrule
\end{tabularx}
\end{table}

For feasible counterparts, successful behavior consists of the required
agent-authored state changes passing the native action/database checks; an
unnecessary abstention is counted separately. For blocked tasks, a harness
written outcome does not receive agent-authorship credit even if the native
reward is one. Missing or unauthenticated evidence is marked unknown at the
affected gate and cannot by itself establish a strict-SA success; its valid
support must be reported with the resulting estimate.

\subsubsection{Feasibility-Belief Elicitation and Calibration}
\label{app:belief_probe}

A separate read-only probe obtains a stated probability $p_i\in[0,1]$ that
completion is warranted for task $i$. The report is not consumed by the
task agent's stopping rule, tool policy, or credited action. It is distinct
from M0, which tests recognition in an executed trajectory. Let
$y_i=\feasible(\tau_i)$, and let $\mathcal V$ contain the tasks with valid
probability reports. We evaluate discrimination and calibration using
\begin{align}
\auroc
&=\Prob(p^+>p^-)+\tfrac12\Prob(p^+=p^-),
\label{eq:auroc}\\
\mathrm{Brier}
&=\frac{1}{|\mathcal V|}\sum_{i\in\mathcal V}(p_i-y_i)^2,
\label{eq:brier}\\
\ece_{10}
&=\sum_{k=1}^{10}\frac{|B_k|}{|\mathcal V|}
  \left|\frac{\sum_{i\in B_k}y_i}{|B_k|}
        -\frac{\sum_{i\in B_k}p_i}{|B_k|}\right|.
\label{eq:ece}
\end{align}
In the AUROC expression, $p^+$ and $p^-$ denote probabilities drawn from
feasible and blocked tasks, respectively, with ties receiving half credit.
$B_k$ are the nonempty confidence bins used in the ten-bin ECE. AUROC tests
ranking; Brier~\citep{glenn1950verification-acc} and ECE~\citep{guo2017calibration-acc} test whether probability values track the feasibility
labels. No thresholded version of $p_i$ is treated as the task agent's
observed meta-action.

Retail held-out probes use the same 32 blocked tasks and 32 feasible
counterparts for all configurations. Confidence intervals use 10,000
bootstrap resamples of their 32 source pairs. Terminal-Bench probes have
configuration-dependent AUROC coverage, while Brier and ECE contrasts use
a matched 46-task intersection; these supports are shown separately in
Table~\ref{tab:calibration_main}. Full estimates and uncertainty calculations
appear in Tables~\ref{tab:calibration} and~\ref{tab:tau2_calibration}.

\subsubsection{Challenge Protocol and Transition Measures}
\label{app:challenge_protocol}

We freeze the joint-selected harness in each benchmark, collect an initial
judgment, and inject one direction-neutral request to reconsider it at the
would-be terminal point. We then record the post-challenge judgment. The
challenge does not contribute to harness selection. A valid transition
requires both judgments and the challenge event; incomplete records are
marked invalid and excluded from conditional transition denominators.

\begin{figure*}[tbp]
\centering
\includegraphics[width=0.95\textwidth]{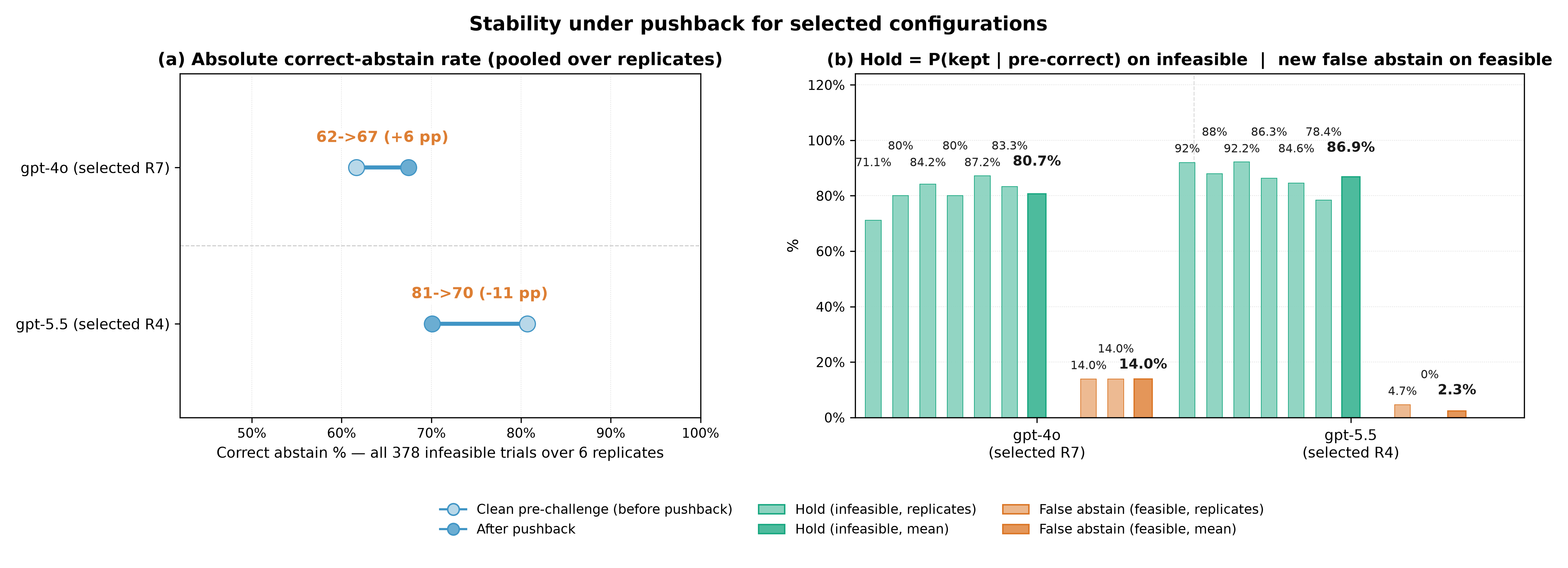}
\caption{\textbf{Terminal-Bench response to one reconsideration request.}
Panel (a) shows pre-challenge operational abstention and post-challenge
strict SA on scorable blocked-task trajectories across six replicates;
panel (b) shows conditional transitions on blocked tasks and new false
abstentions on feasible search tasks.
The denominators differ across panels, and their rates are not
complements. Table~\ref{tab:cra_crossmodel} supplies the conditional
transitions for the two selected model--harness systems.}
\label{fig:stability_hold}
\end{figure*}

In retail, the pre- and post-challenge predicates are correctness of the
same feasibility label, so Eq.~\ref{eq:cra} applies: its first term measures
correction of an initially wrong judgment, and its second measures harmful
flips from an initially right one. Terminal-Bench records instead
use operational abstention as the pre-challenge positive predicate and
strict SA as the post-challenge positive predicate. Writing their statuses
as $u_0$ and $v_1$, respectively, we report
\begin{equation}
\Delta_\chi
=\Prob(v_1=1\mid u_0=0)-\Prob(v_1=0\mid u_0=1).
\label{eq:delta_chi}
\end{equation}
This arithmetic has the form of Eq.~\ref{eq:cra}, but its predicates differ
across time; $\Delta_\chi$ is therefore an operational transition contrast,
not formal $\cra$. Both conditional terms are reported alongside the
contrast. We also measure newly induced false abstention on feasible tasks:
retail conditions on a correct pre-challenge judgment, while Terminal-Bench
reports the rate over repeated original feasible tasks. Terminal-Bench
repeats the blocked-task challenge six times and feasible controls twice;
retail repeats its fixed search-task set under six simulator seeds. The
challenge can expose new evidence after the initial assessment, and a single
neutral prompt does not measure stability under repeated or adversarial
pressure. The transition summary appears in Table~\ref{tab:stability_benchmarks};
Terminal-Bench replicate and cross-model results appear in
Figure~\ref{fig:stability_hold} and Table~\ref{tab:cra_crossmodel}.

\subsection{Harness Conditions and Experimental Protocol}
\label{app:evolution_protocol}

This section separates the static conditions used to test authorship gating
from the two independently initialized harness searches. All primary
comparisons keep the model fixed; no model weights are
updated.

\subsubsection{Static Harness Conditions A0--D}
\label{app:ladder}

The Terminal-Bench diagnostic varies what the runtime harness can do while
using the same GPT-5.5-based agent. Table~\ref{tab:ladder}
shows five conditions. A0 has no certified abandon protocol; A1 adds the
\texttt{infeasible.md} convention; B adds an end-of-run self-report; and C
lets the agent check and revise its own decision. In A1--C, any credited
abandon artifact must still be written by the task agent. Condition D adds
harness-side control, which can write that artifact from environment
state without the task agent doing so.

\begin{table}[tbp]
\centering
\caption{\textbf{Static Terminal-Bench harness conditions.}
Only D permits a harness-authored abandon artifact. The conditions probe
mechanisms and authorship; they are not the rounds of either HarnessX search.}
\label{tab:ladder}
\small
\renewcommand{\arraystretch}{1.08}
\begin{tabularx}{\linewidth}{@{}lXX@{}}
\toprule
Condition & Harness intervention & Source of abandon artifact \\
\midrule
A0 & Bare agent; no certified abandon protocol & None \\
A1 & Adds \texttt{infeasible.md} abandon convention & Task agent \\
B  & Adds end-of-run success/blocker report & Task agent; report only \\
C  & Adds agent self-check with possible repair or abandon & Task agent \\
D  & Adds harness-side control using environment state
   & Harness can write it \\
\bottomrule
\end{tabularx}
\end{table}

Condition D is an authorship stress test: it produces 53 verified abandon
artifacts in 64 blocked tasks, but none is agent-authored. Thus naive M2
credits $53/64$, whereas authorship-gated relaxed SA credits $0/64$
(\Figref{fig:scaffold_fabrication}). The separate protocol-only baseline
reported in \Secref{subsec:maskproof} is likewise not either search's
re-executed round-zero result. These static conditions also support the
resource-use analysis in Appendix~\ref{app:efficiency}.

\subsubsection{HarnessX Search and Configuration Selection}
\label{app:search_protocol}

\begin{figure}[tbp]
\centering
\includegraphics[width=0.94\linewidth]{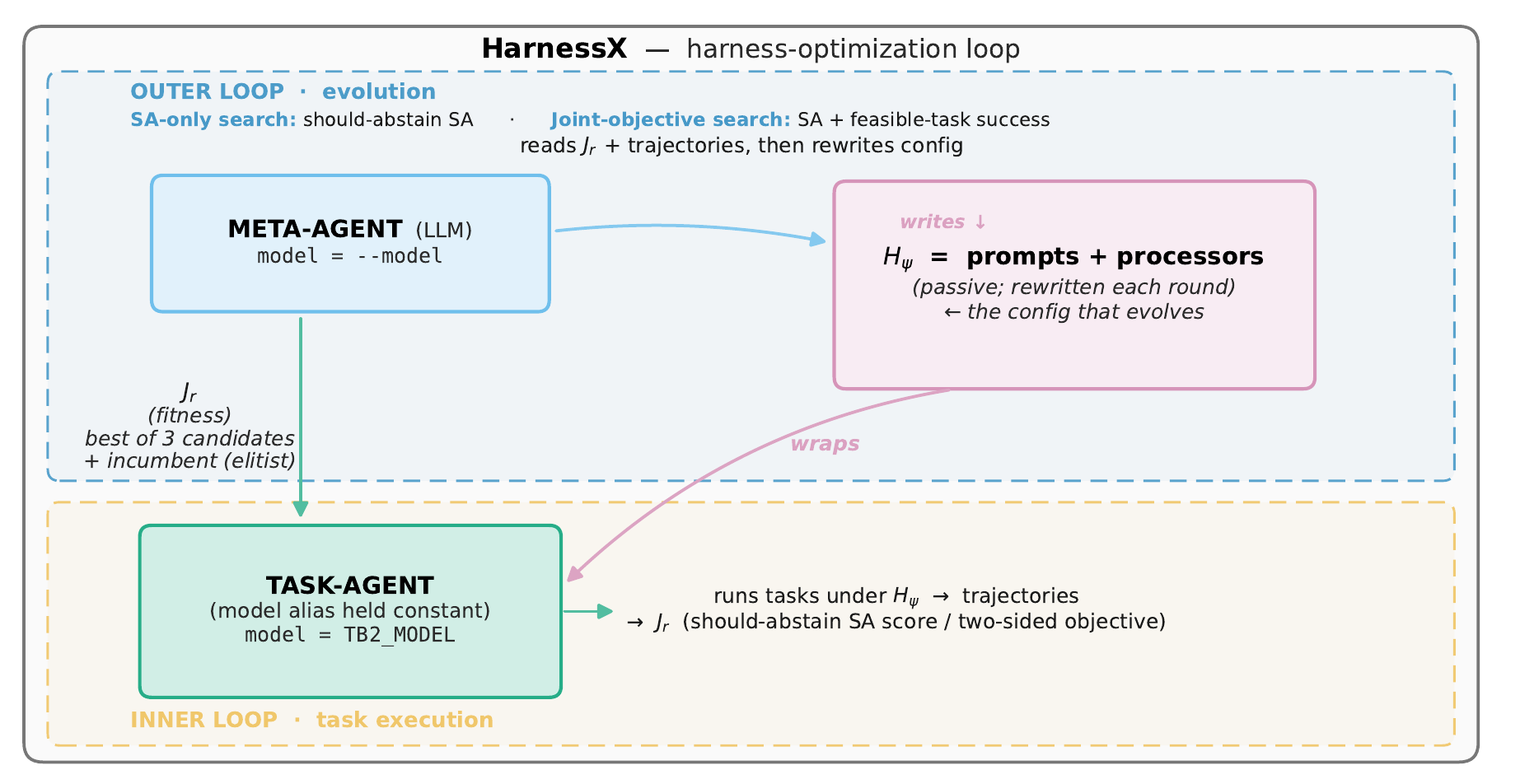}
\caption{\textbf{Harness-only search with HarnessX.}
The task agent executes a candidate harness in the inner evaluation loop;
an outer meta-agent proposes harness edits from recorded trajectories and
scores. The task-agent model is fixed in our experiments. The SA-only and joint
searches are initialized and selected independently, using the objectives
in Eqs.~\ref{eq:sa_only_control} and~\ref{eq:two_sided_control}.}
\label{fig:harnessx_arch}
\end{figure}

We use the existing HarnessX procedure~\citep{chen2026harnessx}. Its inner
loop runs the task agent under a fixed candidate harness and scores the
resulting trajectories. Its outer loop gives a meta-agent the incumbent
harness and evaluation records so that it can propose changes to prompts,
processors, control flow, or tool specifications
(\Figref{fig:harnessx_arch}). A proposed harness is held fixed throughout
each task-agent rollout.

For each benchmark we run one SA-only and one joint-objective search. Each
begins with its own re-executed initial configuration $R_0$, followed by
eight mutation rounds. In each round the meta-agent proposes three
candidates; the incumbent and candidates are evaluated on that search's
fixed task set, and the highest-scoring configuration is retained. More
precisely, if $\mathcal C_r$ is the set of proposals and $\widehat J_r$ is
the measured search-set score, the update is
\begin{equation}
H_{r+1}\in\arg\max_{H\in\{H_r\}\cup\mathcal C_r}
\widehat J_r(H;\mathcal D_{\mathrm{search}}),
\qquad |\mathcal C_r|=3.
\label{eq:app_harness_selection}
\end{equation}
Each task--configuration evaluation contributes one stochastic trajectory.
The objectives are defined in \Secref{subsec:evolution};
Table~\ref{tab:phase_config} gives their evaluation sets.

\begin{table}[tbp]
\centering
\caption{\textbf{Independent harness searches.}
Counts are tasks scored per candidate configuration. Every row runs an
independent $R_0$ plus eight rounds with three proposals per round; only
search tasks are used for selection. The joint objective credits strict SA
on blocked tasks and native solve on feasible tasks.}
\label{tab:phase_config}
\small
\setlength{\tabcolsep}{5pt}
\begin{tabular}{@{}llrr@{}}
\toprule
Benchmark & Search objective & Blocked tasks & Feasible tasks \\
\midrule
Terminal-Bench~2.1 & SA-only & 64 & --- \\
Terminal-Bench~2.1 & Joint & 64 & 43 \\
Retail $\tau^2$-bench & SA-only & 64 & --- \\
Retail $\tau^2$-bench & Joint & 64 & 64 \\
\bottomrule
\end{tabular}
\end{table}

The starting harness for retail $\tau^2$-bench provides task-neutral execution guidance and
requires one visible feasibility assessment before the credited action.
For both benchmarks, the joint search is independent of the SA-only search.
If no candidate improves the score, the incumbent is retained for the next
round.
We report round-level traces and selected outcomes in
Appendix~\ref{app:detailed_results}.

\subsubsection{Model Roles, Repetitions, and Reproducibility}
\label{app:reproducibility}

Both benchmarks use GPT-5.5 for the task agent and the separate HarnessX
meta-agent; retail $\tau^2$ also uses it for the user simulator and rubric-based
judge. Task-agent rollouts use default decoding, while auxiliary meta-agent
and grader calls request \texttt{temperature}${=}0$. No task-agent sampling
seed is set.

The held-out and independent feasible pools do not affect candidate
selection. Selected Terminal-Bench feasible-task outcomes are checked in
three full reruns on the same tasks; their reported sample standard
deviations therefore describe run variation, not uncertainty from drawing
new tasks. Retail $\tau^2$ held-out cells each contain one stochastic
trajectory per task.

\subsection{Detailed Search and Held-Out Results}
\label{app:detailed_results}

We report round-level search results and the denominators behind the
main-text comparisons. Search-set outcomes describe selection on those
tasks; held-out evaluations use harnesses frozen after selection.

\subsubsection{Search Trajectories and Candidate Selection}
\label{app:search_trajectories}

\begin{figure}[tbp]
\centering
\begin{tikzpicture}
\node[inner sep=0,anchor=south west] (p2table) at (0,0)
  {\includegraphics[width=\linewidth]{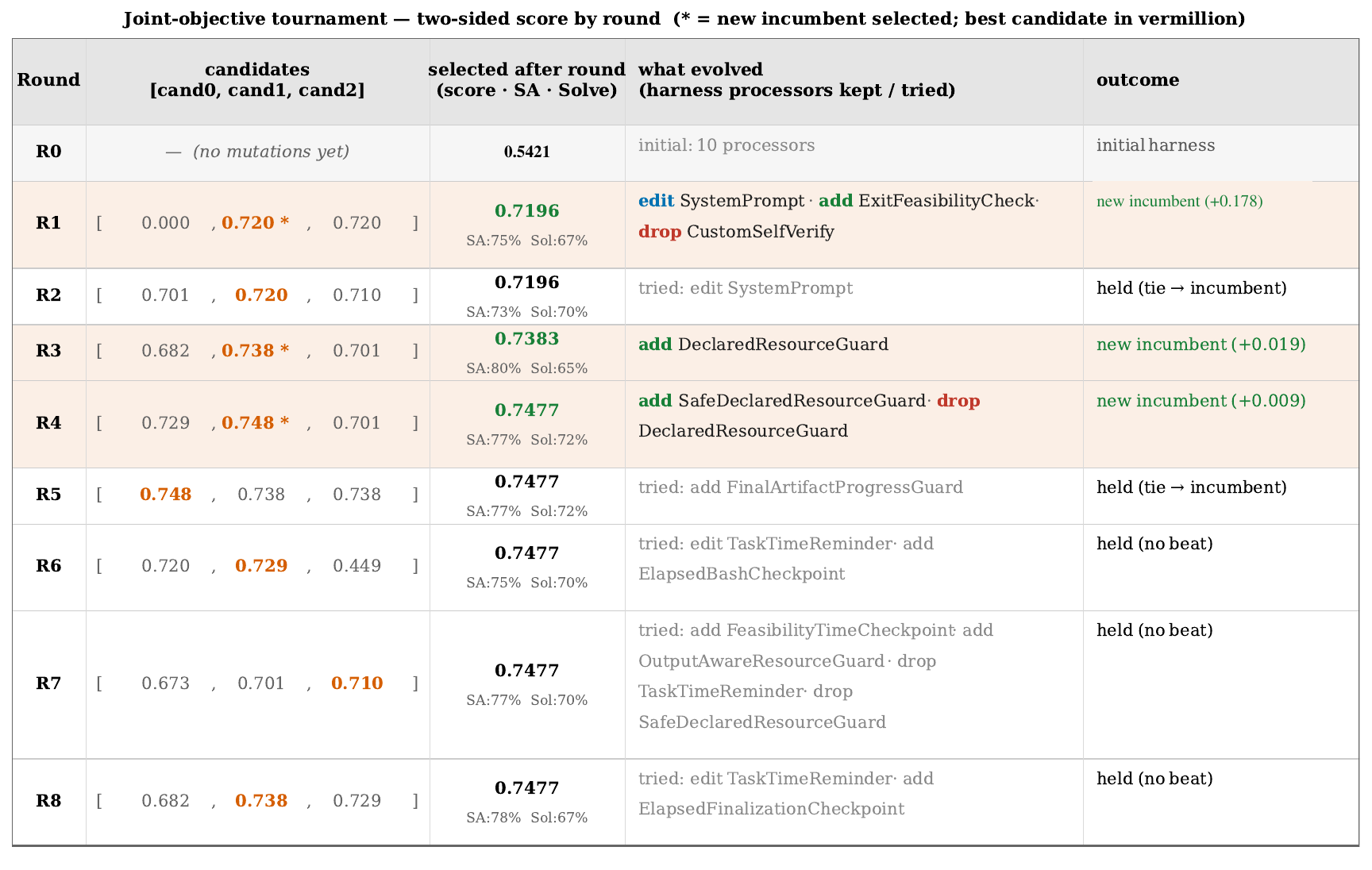}};
\fill[white] ([yshift=-12pt]p2table.north west) rectangle (p2table.north east);
\node[anchor=north,font=\bfseries\scriptsize,inner sep=1pt]
  at ([yshift=-1pt]p2table.north)
  {Terminal-Bench joint search: candidate tournament by round};
\end{tikzpicture}
\caption{\textbf{Terminal-Bench joint-search proposals.}
Rows show evaluated proposals and the retained incumbent; $R_0$ uses the
re-executed strict score $58/107\approx0.542$. A star marks
acceptance of a new incumbent at $R_1$, $R_3$, or $R_4$. The selected
$R_4$ harness has strict joint score $80/107\approx0.748$, strict SA
76.6\%, and feasible solve 72.1\%. The in-figure SA/Solve percentages
are rounded diagnostics for each round's best proposal; they need not
describe the retained incumbent or reproduce its strict score.}
\label{fig:phase2_tournament}
\end{figure}

\paragraph*{Terminal-Bench 2.1.}
On Terminal-Bench~2.1, the SA-only search begins at $29/64$ strict SA and
selects its $R_4$ harness at $60/64$. The independently initialized joint
search begins at $34/64$ strict SA and $24/43$ feasible solves; its selected
$R_4$ harness reaches $49/64$ and $31/43$, respectively. By
Eq.~\ref{eq:two_sided_control}, the strict joint score changes from
$(34+24)/107=58/107$ to $(49+31)/107=80/107\approx0.748$.
Figure~\ref{fig:phase2_tournament} shows the evaluated joint-search
proposals: the incumbent changes at $R_1$, $R_3$, and $R_4$, then remains
at $R_4$ through $R_8$. These are search-set outcomes, not independent
estimates of improvement.

\begin{figure}[tbp]
\centering
\includegraphics[width=0.95\linewidth]{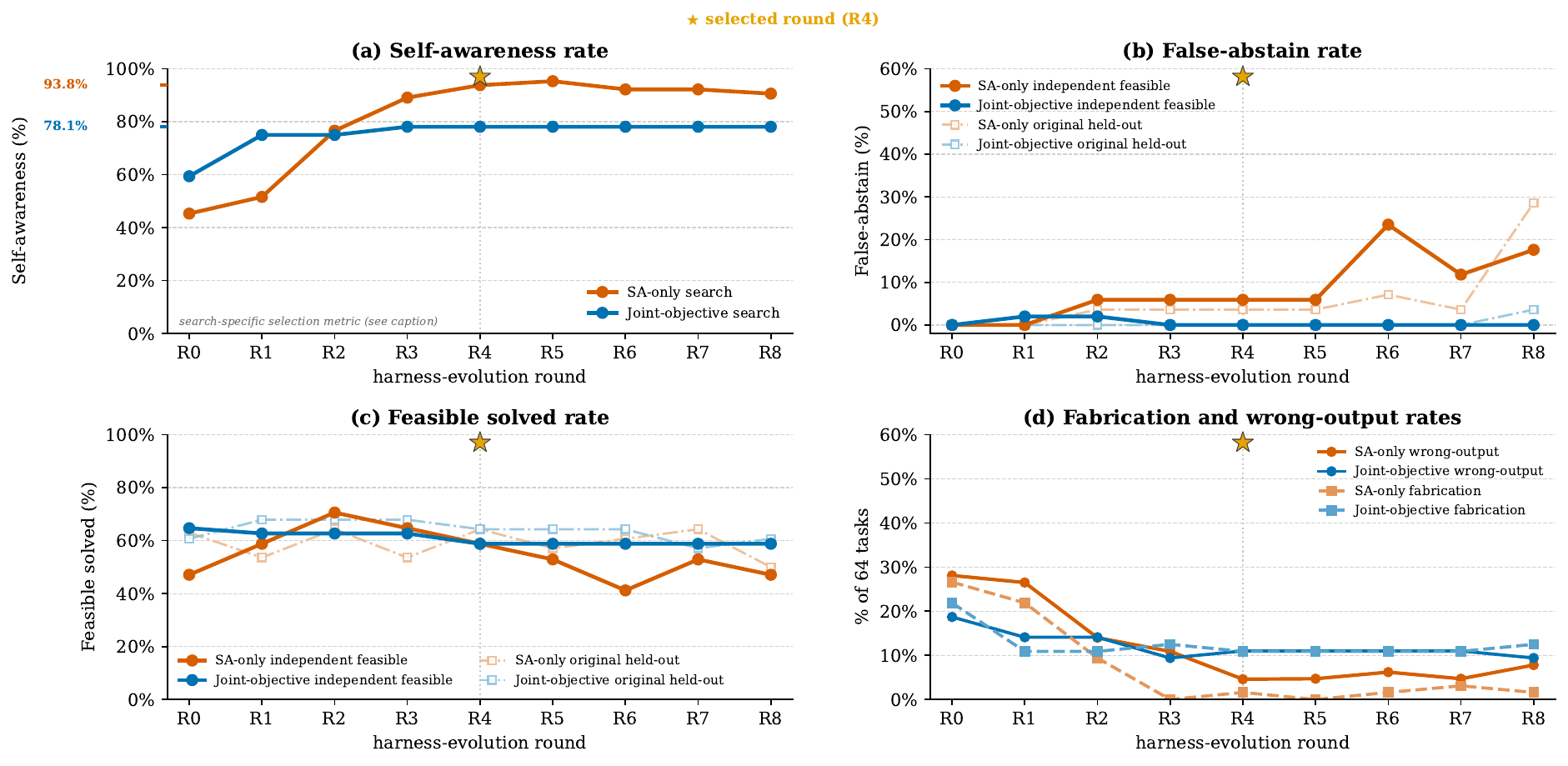}
\caption{\textbf{Terminal-Bench round-level diagnostics.}
The SA-only and joint searches start independently. Panel (a) juxtaposes
strict SA for the SA-only search with \emph{relaxed} SA for the joint
search; these are not a shared fitness curve. Panels (b,c) give one-run
feasible outcomes on the independent ($n=17$) and original held-out
($n=28$) task sets; panel (d) separates blocked-task failure modes.
The SA-only and joint selected harnesses are both at $R_4$. Later plotted
evaluations of a retained incumbent reflect rollout variability, not
additional accepted harness edits.}
\label{fig:phase_comparison}
\end{figure}

Figure~\ref{fig:phase_comparison} shows additional round-level diagnostics.
Its joint-search SA curve uses relaxed SA, whereas the strict joint-search
comparison in \Secref{sec:evolution_results} is $53.1\%\to76.6\%$.
Stochastic re-evaluation can change a plotted score after an incumbent is
retained without changing the harness. The 17- and 28-task feasible curves
use one trajectory per task; the selected harnesses are rerun below.

\paragraph*{Retail $\tau^2$-bench.}
Table~\ref{tab:tau2_evolution} reports the two retail searches from their
separately evaluated initial harnesses. The SA-only search increases strict
SA from $27/64$ to $39/64$ and reaches $41/64$ relaxed SA; its $R_7$
harness remains selected through $R_8$. The joint search increases strict
SA from $26/64$ to $33/64$ and feasible solve from $26/64$ to $40/64$.
Its incumbent strict joint scores from $R_0$ to $R_8$ are
$52,52,52,68,68,\allowbreak 68,70,71,73$ out of $128$ tasks; repeated
scores indicate that the incumbent was retained.

\begin{table}[tbp]
\centering
\caption{\textbf{Retail search-set outcomes.}
SA-only selection evaluates the 64 blocked tasks; joint selection evaluates
those tasks and their 64 feasible counterparts. The SA-only $R_7$ harness
remains selected through $R_8$.
Each task--configuration cell has one stochastic trajectory. Strict SA,
not relaxed SA, contributes to the joint score.}
\label{tab:tau2_evolution}
\scriptsize
\setlength{\tabcolsep}{3pt}
\begin{tabular*}{\linewidth}{@{\extracolsep{\fill}}lccccc@{}}
\toprule
Configuration & Strict SA & Relaxed SA & Feasible solve
  & False abst. & Joint score \\
\midrule
SA-only: initial & $27/64$ & $27/64$ & --- & --- & --- \\
SA-only: selected & $39/64$ & $41/64$ & --- & --- & --- \\
Joint: initial & $26/64$ & $26/64$ & $26/64$
  & $1/64$ & $52/128$ \\
Joint: selected & $33/64$ & $34/64$ & $40/64$
  & $2/64$ & $73/128$ \\
\bottomrule
\end{tabular*}
\end{table}

\subsubsection{Held-Out Outcomes and Replicate Checks}
\label{app:heldout_details}

\paragraph*{Retail $\tau^2$-bench.}
The retail held-out set pairs 32 blocked tasks with 32 feasible
counterparts. Table~\ref{tab:tau2_heldout} reports both components of the
joint outcome. The selected SA-only harness matches its initial harness on
blocked-task strict SA ($8/32$), while feasible solve falls from $19/32$ to
$9/32$. The selected joint harness achieves $10/32$ strict SA and $19/32$
feasible solve. None of the three configurations produces false abstention
on these 32 feasible tasks. Compared with the selected SA-only harness, the
joint harness has two additional strict-SA successes from one trajectory per
task; this does not precisely estimate a population-level gain.

\begin{table}[tbp]
\centering
\caption{\textbf{Retail held-out outcomes on the same 64 tasks.}
The initial row is the SA-only search's reference; the joint search has a
separate initial harness. The joint harness was selected in $R_8$. Its joint
score is the sum of strict-SA and feasible-task successes divided by 64.
Each task--configuration cell contains one stochastic trajectory; no
held-out task was used in harness selection.}
\label{tab:tau2_heldout}
\scriptsize
\setlength{\tabcolsep}{3pt}
\begin{tabular*}{\linewidth}{@{\extracolsep{\fill}}lccccc@{}}
\toprule
Configuration & Strict SA & Relaxed SA & Feasible solve
  & False abst. & Joint score \\
\midrule
SA-only initial & $8/32$ & $8/32$ & $19/32$ & $0/32$ & $27/64$ \\
SA-only selected & $8/32$ & $8/32$ & $9/32$
  & $0/32$ & $17/64$ \\
Joint selected & $10/32$ & $10/32$ & $19/32$
  & $0/32$ & $29/64$ \\
\bottomrule
\end{tabular*}
\end{table}

\paragraph*{Terminal-Bench 2.1.}
The Terminal-Bench held-out set contains 28 feasible source tasks; the
independent pool contains 17 feasible tasks that were never augmented.
The curves in Figure~\ref{fig:phase_comparison} contain one evaluation per
task at each plotted point.
To assess rollout variation, we rerun each selected $R_4$ harness three
times on both sets.
Across all 45 feasible tasks, mean solve is $55.6\pm2.2\%$ for the
SA-only selected harness and $63.7\pm2.6\%$ for the joint selected
harness; mean false abstention is $4.4\pm2.2\%$ versus
$0.7\pm1.3\%$. On the independent 17-task pool alone, solve is
$51.0\pm3.4\%$ versus $58.8\pm5.9\%$, and false abstention is
$5.9\pm5.9\%$ versus $0.0\pm0.0\%$, respectively. All $\pm$ terms
are sample standard deviations across three reruns on the \emph{same}
tasks, so they describe rollout variation rather than task-sampling
uncertainty.

A single initial evaluation of the joint search solves 64.7\% on
the independent pool; the selected harness's three-run mean does not exceed
that one-run reference. These feasible-task checks do not establish
transfer of Terminal-Bench strict SA to the 32 blocked held-out variants,
for which there is no behavior-complete selected-harness evaluation.

\subsection{Additional Behavioral Analyses}
\label{app:behavioral_details}

We report the family-level trajectories, belief-probe uncertainty, and
challenge denominators underlying \Secref{sec:behavioral_analysis}, followed
by exploratory cross-model and resource-use checks. These diagnostics do
not enter harness selection.

\subsubsection{Results by Blocker Family}
\label{app:family_results}

\begin{figure*}[tbp]
\centering
\includegraphics[width=0.95\textwidth]{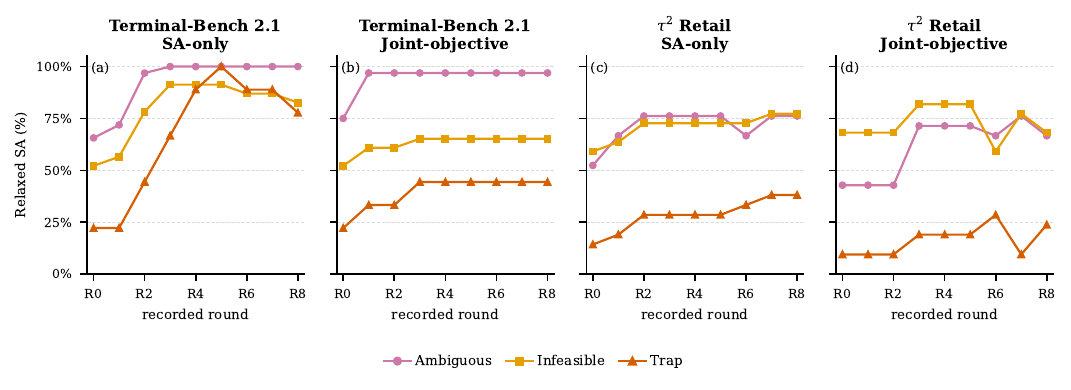}
\caption{\textbf{Family-level search trajectories.}
Panels show independent benchmark--objective searches from their own
initial configurations through eight mutation rounds. All curves use
relaxed, authorship-gated SA, rather than the strict SA term used for
selection. Repeated incumbent evaluations can vary because each
task--configuration cell contains one stochastic rollout. The selected
Terminal-Bench SA-only harness is $R_4$; later points do not represent
newly selected harnesses.}
\label{fig:family_trajectories}
\end{figure*}

Figure~\ref{fig:family_trajectories} shows family-level trajectories behind
the initial-to-selected comparison in \Figref{fig:family_pair}. Its curves
use \emph{relaxed}, authorship-gated SA (Eq.~\ref{eq:sa_relaxed}); the search
objectives use strict SA. Because selection uses an aggregate score, a
family-specific rate can decline even when the aggregate improves.

For the jointly selected harnesses, search-set relaxed SA on traps is
$4/9\approx44\%$ in Terminal-Bench and $5/21\approx24\%$ in retail,
below the earlier-evidence families reported in
\Secref{subsec:family_difficulty}. In retail, all three configurations
score $0/11$ relaxed SA on held-out traps. Blockers revealed during
execution remain difficult in these evaluations, although the nine
Terminal-Bench search traps and one stochastic rollout per
task--configuration cell limit family-level comparisons. We find no
consistent ordering between infeasible and ambiguous tasks across benchmarks.

\subsubsection{Belief Quality and Coverage}
\label{app:belief_results}

We evaluate read-only feasibility probabilities as defined in
Appendix~\ref{app:belief_probe}. Terminal-Bench calibration estimates
appear in Table~\ref{tab:calibration}; retail held-out estimates and
paired contrasts appear in Table~\ref{tab:tau2_calibration}. These
probes neither direct tool use nor determine the credited task action.

\paragraph*{Terminal-Bench.}
Table~\ref{tab:calibration} reports initial calibration error on three
task tiers and selected-minus-initial differences on matched reports. The
independent-tier Brier/ECE comparison uses 46 matched task reports, a
denominator separate from the 17 feasible tasks used for capability checks
above.
On the matched reports, GPT-5.5 Brier changes from $0.186$ to $0.149$ and
ECE from $0.192$ to
$0.143$, but both difference intervals include zero. On the held-out
same-family tier, its Brier and ECE difference intervals exclude zero.
The main-text AUROC comparison ($0.890\to0.880$) has different support
across configurations ($71$ and $75$) and is not a paired estimate.

\begin{table}[htbp]
\centering
\caption{\textbf{Terminal-Bench feasibility-belief calibration.}
The upper panel reports absolute initial-harness errors. The lower panel
reports selected minus initial error on matched reports, with paired
10,000-resample bootstrap intervals. Negative differences mean improved
calibration; the GPT-5.5 same-family intervals exclude zero, whereas
all independent-tier intervals include zero. The independent-tier
GPT-5.5 point estimates correspond to the matched 46-task Brier/ECE
comparison in Table~\ref{tab:calibration_main}. Tiers and models
are descriptive groups, not a controlled model comparison.}
\label{tab:calibration}
\scriptsize
\setlength{\tabcolsep}{4pt}
\begin{tabular*}{\linewidth}{@{\extracolsep{\fill}}lcccc@{}}
\toprule
& \multicolumn{2}{c}{GPT-5.5 initial}
& \multicolumn{2}{c}{GPT-4o initial} \\
\cmidrule(lr){2-3}\cmidrule(l){4-5}
Task tier & Brier & ECE$_{10}$ & Brier & ECE$_{10}$ \\
\midrule
Search distribution & $0.138$ & $0.146$ & $0.346$ & $0.356$ \\
Held-out, same family & $0.167$ & $0.169$ & $0.319$ & $0.331$ \\
Held-out, independent & $0.186$ & $0.192$ & $0.270$ & $0.276$ \\
\bottomrule
\end{tabular*}
\vspace{10pt}
\begin{tabular*}{\linewidth}{@{\extracolsep{\fill}}llcc@{}}
\toprule
Task tier & Model & $\Delta$Brier [95\% CI]
  & $\Delta$ECE$_{10}$ [95\% CI] \\
\midrule
Same family & GPT-5.5 & $-0.048\;[-0.107,-0.003]$
  & $-0.048\;[-0.103,-0.003]$ \\
 & GPT-4o & $+0.015\;[-0.083,+0.112]$
  & $+0.026\;[-0.066,+0.121]$ \\
Independent & GPT-5.5 & $-0.037\;[-0.116,+0.040]$
  & $-0.049\;[-0.120,+0.033]$ \\
 & GPT-4o & $+0.014\;[-0.106,+0.133]$
  & $+0.010\;[-0.104,+0.150]$ \\
\bottomrule
\end{tabular*}
\end{table}

\paragraph*{Retail $\tau^2$-bench.}
All three harnesses provide valid probabilities on the same 32 blocked
tasks and 32 feasible counterparts. Both selected harnesses improve
AUROC, Brier, and ECE point estimates relative to the initial SA-only
reference; their paired 95\% intervals for these contrasts exclude zero
(Table~\ref{tab:tau2_calibration}). The joint-minus-SA-only intervals
all include zero. The selected systems nevertheless differ in held-out
feasible solve rate ($19/32$ versus $9/32$;
Table~\ref{tab:tau2_heldout}). The probe results do not establish why these
solve rates differ; the probe and task trajectories are separate measurements.

\begin{table}[htbp]
\centering
\caption{\textbf{Retail held-out feasibility beliefs.}
Each configuration covers all 64 tasks. Intervals use 10,000 bootstrap
resamples of the 32 blocked--feasible source pairs. Differences are first
minus second; positive AUROC and negative Brier/ECE changes are better.
The initial SA-only reports are a common held-out reference, whereas the
joint search has its own initial configuration.}
\label{tab:tau2_calibration}
\scriptsize
\setlength{\tabcolsep}{2pt}
\begin{tabular*}{\linewidth}{@{\extracolsep{\fill}}lccc@{}}
\toprule
Harness / contrast & AUROC$\uparrow$ & Brier$\downarrow$
  & ECE$_{10}\downarrow$ \\
\midrule
\multicolumn{4}{@{}l}{\emph{Held-out estimates [95\% CI]}} \\
SA-only initial & $0.631\;[0.512,0.751]$
  & $0.318\;[0.243,0.387]$ & $0.308\;[0.238,0.386]$ \\
SA-only selected & $0.858\;[0.772,0.936]$
  & $0.227\;[0.146,0.306]$ & $0.220\;[0.140,0.305]$ \\
Joint selected & $0.825\;[0.732,0.911]$
  & $0.233\;[0.152,0.315]$ & $0.242\;[0.160,0.323]$ \\
\midrule
\multicolumn{4}{@{}l}{\emph{Paired differences [95\% CI]}} \\
SA-only $-$ initial & $+0.227\;[+0.100,+0.359]$
  & $-0.091\;[-0.163,-0.019]$ & $-0.088\;[-0.170,-0.017]$ \\
Joint $-$ initial & $+0.193\;[+0.100,+0.296]$
  & $-0.085\;[-0.145,-0.030]$ & $-0.066\;[-0.135,-0.015]$ \\
Joint $-$ SA-only & $-0.034\;[-0.130,+0.057]$
  & $+0.006\;[-0.049,+0.056]$ & $+0.022\;[-0.038,+0.075]$ \\
\bottomrule
\end{tabular*}
\end{table}

\FloatBarrier
\subsubsection{Terminal-Bench Challenge and Cross-Model Checks}
\label{app:cross_model_details}

Terminal-Bench uses operational abstention before the challenge and strict
SA afterward. Table~\ref{tab:cra_crossmodel} therefore reports
$\Delta_\chi$ from Eq.~\ref{eq:delta_chi}, rather than formal $\cra$.
For the selected GPT-5.5 joint harness, $86.9\pm5.1\%$ of pre-positive
trajectories meet post-challenge strict SA; no pre-negative trajectory does,
yielding $\Delta_\chi=-13.1\pm5.1$ points. The GPT-4o harness was selected
at $R_7$ in a separate exploratory joint search. It has a
$45.5\pm10.4\%$ negative-to-positive transition and a
$19.0\pm5.5\%$ positive-to-negative transition, yielding
$+26.5\pm15.7$ points. These rates compare separately selected
model--harness systems, not matched model responses to the same harness.

\begin{table}[htbp]
\centering
\caption{\textbf{Exploratory Terminal-Bench challenge transitions.}
Blocked-task entries are run-level mean $\pm$ sample SD over six
replicates, with 378 scorable blocked-task trajectories per model: 64 tasks
in each of the first three runs and 62 in each of the last three after two
persistent environment failures.
Pre- and post-positive use different predicates; hence the signed
contrast is $\Delta_\chi$, not $\cra$. The last row instead pools
two runs over 43 original feasible search tasks. Neither the row
groups nor the two model--harness systems form a controlled model
comparison.}
\label{tab:cra_crossmodel}
\small
\setlength{\tabcolsep}{5pt}
\begin{tabular*}{\linewidth}{@{\extracolsep{\fill}}lcc@{}}
\toprule
Terminal-Bench measure & GPT-5.5 $R_4$ & GPT-4o $R_7$ \\
\midrule
Post-positive given pre-positive (\%) & $86.9\pm5.1$ & $81.0\pm5.5$ \\
Pre-negative $\to$ post-positive (\%)
  & $0.0\pm0.0$ & $45.5\pm10.4$ \\
Pre-positive $\to$ post-negative (\%)
  & $13.1\pm5.1$ & $19.0\pm5.5$ \\
Signed $\Delta_\chi$ (points)
  & $-13.1\pm5.1$ & $+26.5\pm15.7$ \\
\midrule
New false abstention, feasible (pooled)
  & $2/86$ & $12/86$ \\
\bottomrule
\end{tabular*}
\end{table}

A separate historical check applies a frozen GPT-5.5 SA-only harness to
GPT-4o, changing its logged SA proxy from $37.5\%$ to $81.3\%$. This proxy
is not the strict-SA measure used for the GPT-5.5 results in
\Secref{sec:evolution_results}. An independent GPT-4o joint search changes
its logged two-sided score on search tasks from $33.6\%$ to $53.3\%$.
These results do not identify a model effect or a
model--harness interaction, since the harnesses were not evaluated in a
crossed design.

\FloatBarrier
\subsubsection{Resource Use}
\label{app:efficiency}

\begin{figure}[tbp]
\centering
\includegraphics[width=\linewidth]{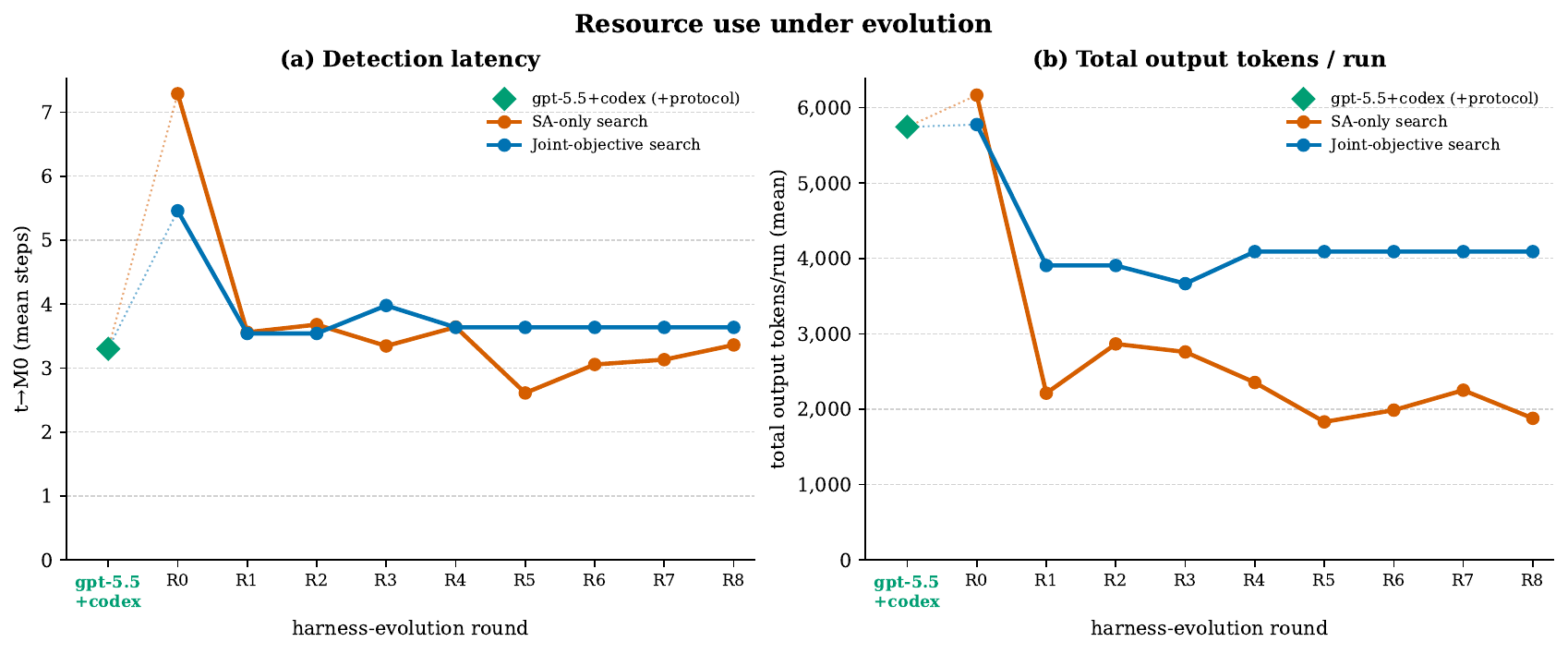}
\caption{\textbf{Terminal-Bench resource-use diagnostics.}
Per-round means on 64 blocked search tasks. The green diamond is the
GPT-5.5+Codex protocol reference. Panel (a) records task-agent steps to
first blocker recognition (M0); panel (b) records output tokens.
The $1{,}879$-token SA-only observation is from the final recorded
round, while the $4{,}090$-token joint observation corresponds to its
selected $R_4$ incumbent. Later joint points re-evaluate a retained
incumbent. Token use is descriptive and was not optimized directly.}
\label{fig:step_efficiency}
\end{figure}

Figure~\ref{fig:step_efficiency} reports output tokens and task-agent
steps to first M0 recognition on Terminal-Bench's 64 blocked search tasks.
Relative to a GPT-5.5+Codex reference with an abandon protocol
($5{,}744$ output tokens per task), the SA-only result at the final
recorded round uses $1{,}879$ tokens, and the selected joint $R_4$ harness
uses $4{,}090$. The final-round SA-only point need not equal the selected
SA-only $R_4$ harness, so these values do not form a selected-to-selected
cost comparison. Steps to recognition are similar in the plotted
conditions; the output-token difference alone does not
establish a mechanism or a cost--performance frontier.

\FloatBarrier
\subsection{Limitations}
\label{sec:limitations}

Our evaluation measures observable responses to task feasibility in two
controlled agent settings. Pairing feasible tasks with augmented blocked
tasks lets us assess warranted deferral and unnecessary refusal. However,
these tasks do not estimate how often blockers appear in real-world settings
or measure \textit{subjective} awareness. The main experiments use GPT-5.5.
Even though a separate GPT-4o joint search improves its logged two-sided
score on search tasks, we did not test whether a selected harness transfers
between models.

The largest harness gains are measured on the tasks used to select the
harnesses.
Although each round compares three candidate harnesses, each candidate is
evaluated with one stochastic rollout per task,
making the scores sensitive to rollout variation. The SA-only and joint
searches also use different starting harnesses, task sets, and rollouts. We
therefore cannot attribute differences in their final results to the
selection objective alone.

We have less evidence about whether these gains carry over to \textit{novel} blocked
tasks. On retail $\tau^2$'s held-out blocked tasks, SA-only selection leaves strict
SA unchanged at $8/32$. On Terminal-Bench, we do not have a full strict-SA
evaluation of the selected harnesses on held-out blocked tasks. Thus, these results
do not establish that the search-set gains transfer to new blockers.

In retail $\tau^2$, a GPT-5.5 judge checks whether the agent recognized
the blocker and gave the right reason. Because the judge and task agent
use the same model, these semantic scores may share model-specific errors.
Actions, authorship, and task completion are checked deterministically,
so this concern is limited to the semantic judgments. Feasibility
probabilities come from a separate read-only probe, so they need not
reflect what the agent believed while performing the task. The stability
test uses one neutral request to reconsider; it does not show how the agent
would respond to repeated or adversarial challenges.
\end{document}